\documentclass[11pt,a4paper]{article}
\usepackage[hyperref]{emnlp-ijcnlp-2019}
\usepackage{microtype}
\usepackage{subfigure}
\usepackage{latexsym}
\usepackage{amssymb}
\usepackage{amsmath}
\usepackage{graphicx}
\usepackage{multirow}
\usepackage{footnote}
\usepackage{times}
\usepackage{url}
\usepackage{textcomp}
\usepackage{CJKutf8}
\usepackage{xcolor}
\usepackage[hang,flushmargin]{footmisc}  

\aclfinalcopy 

\makesavenoteenv{tabular}
\newcommand{\BS}[0]{\textbackslash}
\newcommand{\textsec}[1]{\textsection\ref{#1}}
\newcommand{\CN}[1]{\begin{CJK}{UTF8}{gbsn}#1\end{CJK}}
\newcommand{\FBOX}[1]{\setlength{\fboxsep}{0pt}\fcolorbox{lightgray}{white}{#1}}

\title{Establishing Strong Baselines for the New Decade:\\Sequence Tagging, Syntactic and Semantic Parsing with BERT}

\author{Han He \\
  Computer Science \\
  Emory University \\
  Atlanta GA 30322, USA \\
  \texttt{han.he@emory.edu} \\\And
  Jinho D. Choi \\
  Computer Science \\
  Emory University \\
  Atlanta GA 30322, USA \\
  \texttt{jinho.choi@emory.edu} \\}

\date{}

\begin{document}
\maketitle

\begin{abstract}
This paper presents new state-of-the-art models for three tasks, part-of-speech tagging, syntactic parsing, and semantic parsing, using the 
cutting-edge contextualized embedding framework known as BERT.
For each task, we first replicate and simplify the current state-of-the-art approach to enhance its model efficiency.
We then evaluate our simplified approaches on those three tasks using token embeddings generated by BERT.
12 datasets in both English and Chinese are used for our experiments.
The BERT models outperform the previously best-performing models by 2.5\% on average (7.5\% for the most significant case).
Moreover, an in-depth analysis on the impact of BERT embeddings is provided using self-attention, which helps understanding in this rich yet representation.
All models and source codes are available in public so that researchers can improve upon and utilize them to establish strong baselines for the next decade. 
\end{abstract}


\section{Introduction}
\label{sec:introduction}

It is no exaggeration to say that word embeddings trained by vector-based language models~\cite{mikolov2013distributed,pennington2014glove,bojanowski:17a} have changed the game of NLP once and for all.
These pre-trained word embeddings trained on large corpus improve downstream tasks by encoding rich word semantics into vector space.
However, word senses are ignored in these earlier approaches such that a unique vector is assigned to each word, neglecting polysemy from the context.

Recently, contextualized embedding approaches emerge with advanced techniques to dynamically generate word embeddings from different contexts.
To address polysemous words, \citet{peters:18a} introduce ELMo, which is a word-level Bi-LSTM language model.
\citet{akbik:18a} apply a similar approach to the character-level, called Flair, while concatenating the hidden states corresponding to the first and the last characters of each word to build the embedding of that word.
Apart from these unidirectional recurrent language models, \citet{devlin:18a} replace the transformer decoder from \citet{radford2018improving} with a bidirectional transformer encoder, then train the BERT system on 3.3B word corpus.
After scaling the model size to hundreds of millions parameters, BERT brings markedly huge improvement to a wide range of tasks without substantial task-specific modifications.

In this paper, we verify the effectiveness and conciseness of BERT by first generating token-level embeddings from it, then integrating them to task-oriented yet efficient model structures (Section~\ref{sec:approach}).
With careful investigation and engineering, our simplified models significantly outperform many of the previous state-of-the-art models, achieving the highest scores for 11 out of 12 datasets (Section~\ref{sec:experiments}).

To reveal the essence of BERT in these tasks, we analyze our tagging models with self-attention, and find that BERT embeddings capture contextual information better than pre-trained embeddings, but not necessarily better than embeddings generated by a character-level language model (Section~\ref{ssec:attention-analysis}). 
Furthermore, an extensive comparison between our baseline and BERT models shows that BERT models handle long sentences robustly (Section~\ref{ssec:distance-analysis}).
One of the key findings is that BERT embeddings are much more related to semantic than syntactic (Section~\ref{ssec:label-analysis}).
Our findings are consistent with the training procedure of BERT, which provides guiding references for future research.

To the best of our knowledge, it is the first work that tightly integrates BERT embeddings to these three downstream tasks and present such high performance.
All our resources including the models and the source codes are publicly available.\footnote{\url{https://github.com/emorynlp/bert-2019}}


%
\cleardoublepage
\section{Related Work}
\label{sec:related-work}

\subsection{Representation Learning}

Rich initial word encodings substantially improve the performance of downstream NLP tasks, which have been studied over decades. Except for matrix factorization methods \citep{pennington2014glove}, most work train language models to predict some words given their contexts. Among these work, CBOW and Skip-Gram \cite{mikolov2013distributed} are pioneers of neural language models extracting features within a fixed length window. Then, \citet{joulin2017bag} augment these models with subword information to handle out-of-vocabulary words. 

To learn contextualized representations, \citet{peters:18a} apply bidirectional language model (bi-LM) to tokenized unlabeled corpus. Similarly, the contextual string embeddings \cite{akbik:18a} model language on character level, which can efficiently extract morphological features. However, bi-LM consists of two unidirectional LMs without left or right context, leading to potential bias on one side. To address this limitation, BERT \cite{devlin:18a} employ masked LM to jointly condition on both left and right contexts, showing impressive improvement in various tasks. 


\subsection{Sequence Tagging}

Sequence tagging is one of the most well-studied NLP tasks, which can be directly applied to part-of-speech tagging (POS) and named entity recognition (NER). 
As a general trend, fine grained features often result in better performance. \citet{akbik:18a} feed the contextual string embeddings into a Bi-LSTM-CRF tagger \citep{huang:15a}, improving tagging accuracy with rich morphological and contextual information. 
In a more meticulously designed system, \citet{bohnet:18a} generate representations from both string and token based character Bi-LSTM language models, then employ a meta-BiLSTM to integrate them.

Besides, joint learning and semi-supervised learning can lead to more generalization.
As a highly end-to-end approach, the character level transition system proposed by \citet{kurita:17a} benefits from joint learning on Chinese word segmentation, POS tagging and dependency parsing.
Recently, \citet{clark:18a} exploit large scale unlabeled data with Cross-View Training (CVT), which improves the RNN feature detector shared between the full model and auxiliary modules.

\subsection{Syntactic Parsing}

Dependency tree and constituency structure are two closely related syntactic forms.
\citet{choe:16a} cast constituency parsing as language modeling, achieving high UAS after conversion to dependency tree.
\citet{kuncoro:17a} investigate recurrent neural network grammars through ablations and gated attention mechanism, finding that lexical heads are crucial in phrasal representation.

Recently graph-based parsers resurge due to their ability to exploit modern GPU parallelization. 
\citet{dozat:17a} successfully implement a graph-based dependency parser with biaffine attention mechanism, showing impressive performance and decent simplicity. \citet{clark:18a} improve the feature detector of the biaffine parser through CVT and joint learning. While \citet{ma:18a} introduce stack-pointer networks to model parsing history of a transition-based parser, with biaffine attention mechanism built-in.

\subsection{Semantic Parsing}

Currently, parsing community are shifting from syntactic dependency tree parsing to semantic dependency graph parsing (SDP). 
As graph nodes can have multi-head or zero head, it allows for more flexible representations of sentence meanings. 
\citet{wang:18a} modify the preconditions of List-Based Arc-Eager transition system \cite{choi:13a}, implementing it with Bi-LSTM Subtraction and Tree-LSTM for feature extraction. 

Among graph-based approaches, 
\citet{peng:17a} investigate higher-order structures across different graph formalisms  with tensor scoring strategy, benefiting from multitask learning.
\citet{dozat:18a} replace the softmax cross-entropy in the biaffine parser with sigmoid cross-entropy, successfully turning the syntactic tree parser into a simple yet accurate semantic graph parser. 
\section{Approach}
\label{sec:approach}


\subsection{Token-level Embeddings with BERT}
\label{ssec:bert-token}

BERT splits each token into subwords using WordPiece~\cite{wu2016google}, which do not necessarily reflect any morphology in linguistics.
For example, `Rainwater' gets split into `Rain' and `\#\#water', while words such as \textit{running} or \textit{rapidly} remain unchanged although typical morphology would split them into \textit{run+ing} and \textit{rapid+ly}.
To obtain token-level embeddings for tagging and parsing tasks, the following two methods are experimented:
\cleardoublepage


\paragraph{Last Embedding} Since the subwords from each token are trained to predict one another during language modeling, their embeddings must be correlated.
Thus, one way is to pick the embedding of the last subword as a representation of the token.

\paragraph{Average Embedding} For a compound word like `doghouse' that gets split into `dog' and `\#\#house', the last subword does not necessarily convey the key meaning of the token. Hence, another way is to take the average embedding of the subwords. 

\begin{table}[htbp!]
\centering\small\resizebox{\columnwidth}{!}{
\begin{tabular}{l||c|c}
\multicolumn{1}{c|}{\bf Model} & \multicolumn{1}{c|}{\bf In-domain} & \multicolumn{1}{c}{\bf Out-of-domain} \\
\hline\hline
BERT$_{\tt BASE}$: \textsc{Last}      &     86.7 &     79.5 \\
BERT$_{\tt BASE}$: \textsc{Average}   &     86.7 & \bf 79.8 \\
\hline
BERT$_{\tt LARGE}$: \textsc{Last}     & \bf 86.8 &     79.4 \\
BERT$_{\tt LARGE}$: \textsc{Average}  &     86.4 & \bf 79.5
\end{tabular}}
\caption{Results from the PSD semantic parsing task (\textsec{ssec:semantic-parsing}) using the last and average embedding methods.}
\label{tbl:bert-last-avg}
\end{table}

\noindent Table~\ref{tbl:bert-last-avg} shows results from a semantic parsing task, PSD (Section~\ref{ssec:semantic-parsing}), using the last and average embedding methods with BERT$_{\tt BASE}$ and BERT$_{\tt LARGE}$ models.\footnote{BERT$_{\tt BASE}$ uses 12 layers, 768 hidden cells, 12 attention heads, and 110M parameters, while BERT$_{\tt LARGE}$ uses 24 layers, 1024 hidden cells, 16 attention heads, and 340M parameters. Both models are uncased, since they are reported to achieve high scores for all tasks except for NER \cite{devlin:18a}.}
The average method is chosen for all our experiments since it gives a marginal advantage to the out-of-domain dataset.



\subsection{Input Embeddings with BERT}
\label{ssec:bert-rnn}

While \citet{devlin:18a} report that adding just an additional output layer to the BERT encoder can build powerful models in a wide range of tasks, its computational cost is too high.
Thus, we separate the BERT architecture from downstream models, and feed pre-generated BERT embeddings, $\mathbf{e}^\text{BERT}$, as  input to task-specific encoders:
$$
\mathcal{F}^\text{i} = \mathrm{Encoder}\left( \mathbf{X} \oplus \mathbf{e}^\text{BERT} \right)
$$

\noindent Alternatively, BERT embeddings can be concatenated with the output of a certain hidden layer: 
$$
\mathcal{F}^\text{h} = \mathrm{Encoder}_\text{[h:]}\left( \mathrm{Encoder}_\text{[:h]}\left( \mathbf{X} \right) \oplus \mathbf{e}^\text{BERT} \right)
$$

\noindent Table~\ref{tbl:bert-input-rnn} shows results from the PSD semantic parsing task (Section~\ref{ssec:semantic-parsing}) using the average method from Section~\ref{ssec:bert-token}.
$\mathcal{F}^\text{i}$ shows a slight advantage for both BERT$_{\tt BASE}$ and BERT$_{\tt LARGE}$ over $\mathcal{F}^\text{h}$; thus, it is chosen for all our experiments.

\begin{table}[htbp!]
\centering\small
\begin{tabular}{l||c|c}
\multicolumn{1}{c|}{\bf Model} & \multicolumn{1}{c|}{\bf In-domain} & \multicolumn{1}{c}{\bf Out-of-domain} \\
\hline\hline
BERT$_{\tt BASE}$: $\mathcal{F}^\text{i}$   & \bf 86.7 & \bf 79.8 \\
BERT$_{\tt BASE}$: $\mathcal{F}^\text{h}$   &     86.5 &     79.5 \\
\hline
BERT$_{\tt LARGE}$: $\mathcal{F}^\text{i}$  & \bf 86.4 & \bf 79.5 \\
BERT$_{\tt LARGE}$: $\mathcal{F}^\text{h}$  &     85.9 &     79.1
\end{tabular}
\caption{Results from the PSD semantic parsing task (Section~\ref{ssec:semantic-parsing}) using $\mathcal{F}^\text{i}$ and $\mathcal{F}^\text{h}$.}
\label{tbl:bert-input-rnn}
\end{table}

\subsection{Bi-LSTM-CRF for Tagging}
\label{ssec:bert-bilstm-crf}

\begin{figure}[htbp!]
\centering
\textbf{\includegraphics[width=\columnwidth]{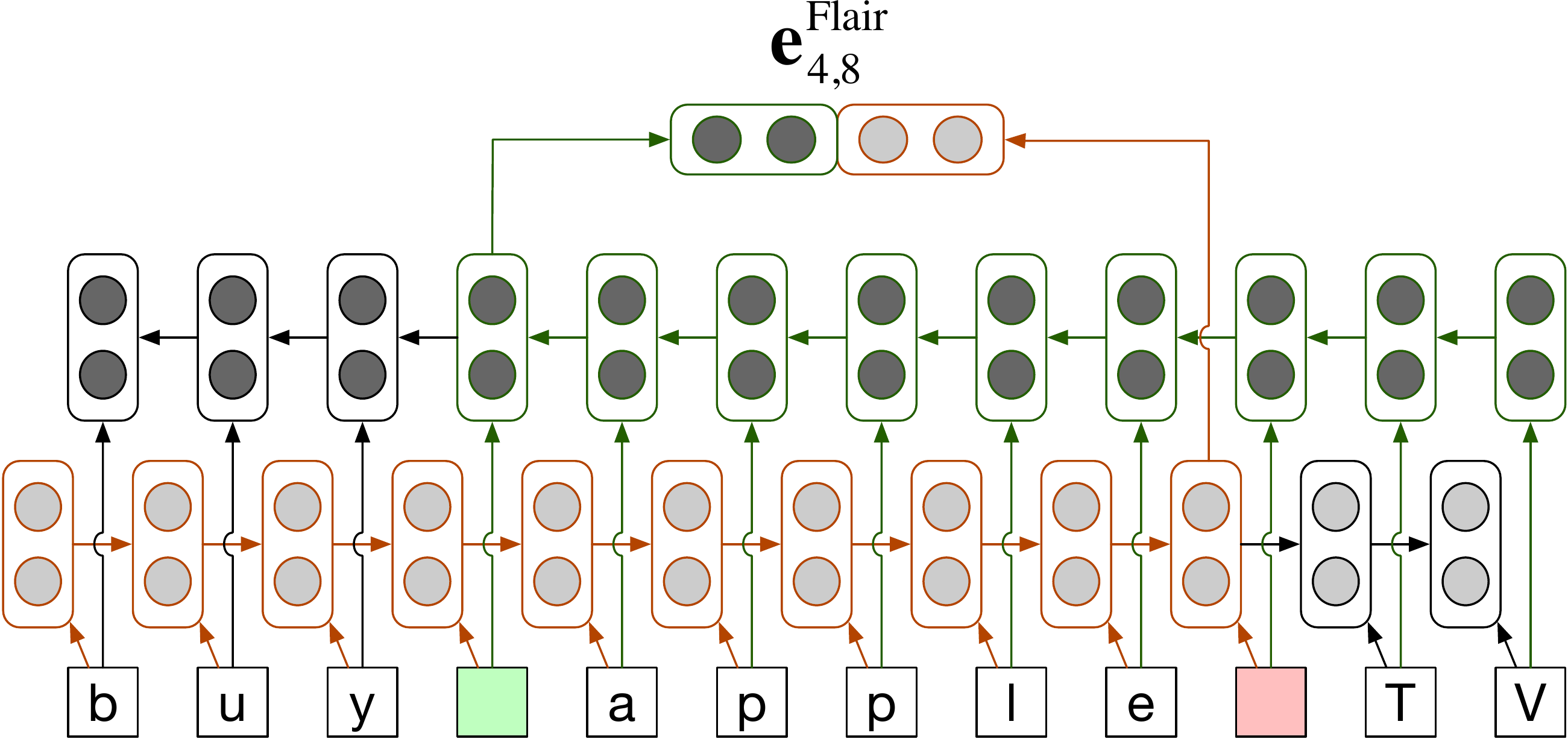}}
\caption{Generating the Flair embedding for `apple'.}
\label{fig:flair}
\end{figure}

For sequence tagging, the Bi-LSTM-CRF~\cite{huang:15a} with the Flair contextual embeddings \cite{akbik:18a}, is used to establish a baseline for English.
Given a token $w$ in a sequence where $c_i$ and $c_j$ are the starting and ending characters of~$w$ ($i$ and $j$ are the character offsets; $i\le j$), the Flair embedding of $w$ is generated by concatenating two hidden states of $c_{j+1}$ from the forward LSTM and $c_{i-1}$ from the backward LSTM (Figure~\ref{fig:flair}):
$$
\mathbf{e}^\text{Flair}_{i,j}={\mathbf{h}}^{\mathrm{f}}(c_{j+1}) \oplus {\mathbf{h}}^{\mathrm{b}}({c_{i-1}})
$$


\noindent $\mathbf{e}^\text{Flair}_{i,j}$ is then concatenated with a pre-trained token embedding of $w$ and fed into the Bi-LSTM-CRF.
In our approach, we present two models, one substituting the Flair and pre-trained embeddings with BERT, and the other concatenating BERT to the other embeddings.
Note that variational dropout is not used in our approach to reduce complexity.

\noindent As Chinese is characterized as a morphologically poor language, the Flair embeddings are not used for tagging tasks; only pre-trained and BERT embeddings are used for our experiments in Chinese.


\begin{figure*}[htbp!]
\centering
\includegraphics[width=\textwidth]{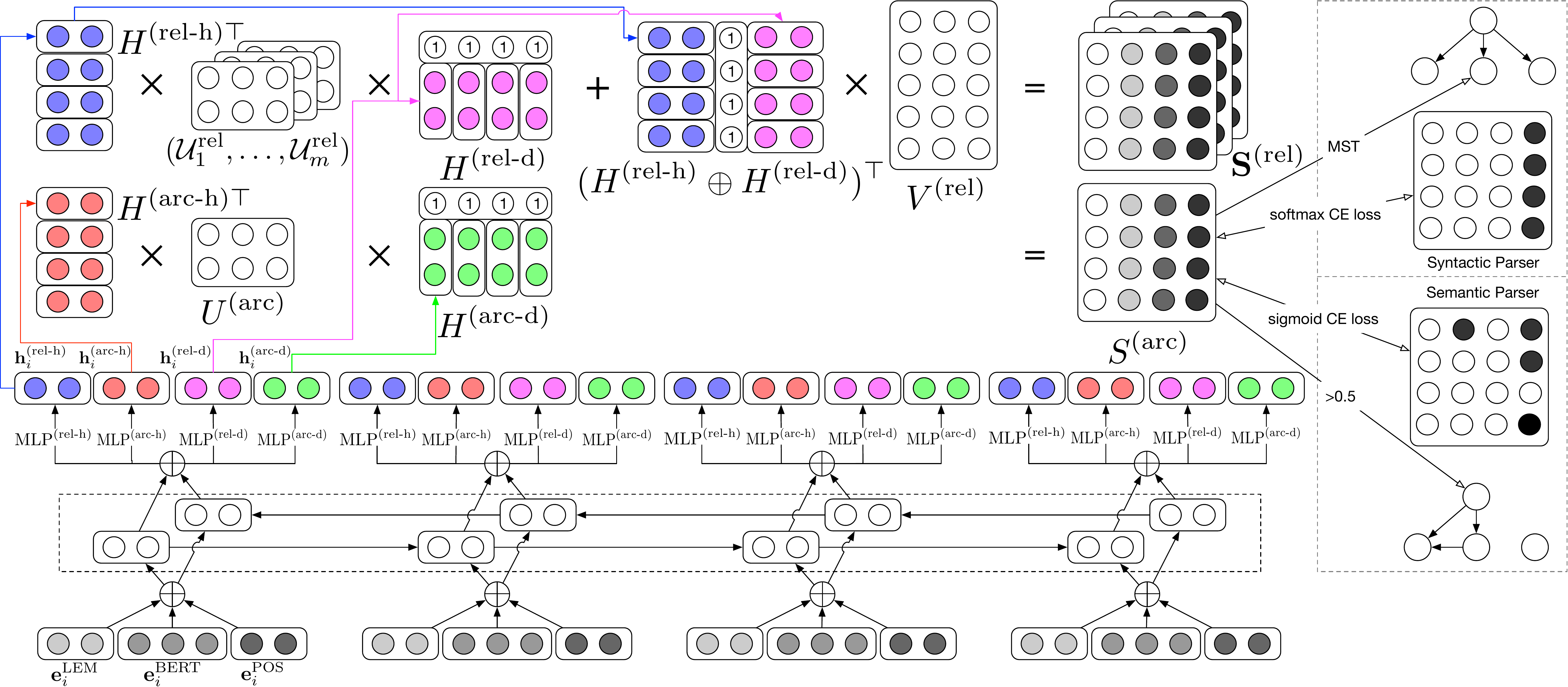}
\caption{Biaffine attention parser}
\label{fig:biaffine}
\end{figure*}

\subsection{Biaffine Attention for Syntactic Parsing}
\label{ssec:bert-biaffine}

A simplified variant of the biaffine parser~\cite{dozat:17a} is used for syntactic parsing (Figure~\ref{fig:biaffine}).
Compared to the original version, the trainable word embeddings are removed and lemmas are used instead of forms to retrieve pre-trained embeddings in our version, leading to less complexity yet better generalization.
\linebreak\noindent Given the $i$'th token $w_i$, the feature vector is created by concatenating its pre-trained lemma embedding $\mathbf{e}^\text{LEM}_i$, POS embedding $\mathbf{e}^\text{POS}_i$ learned during training and the representation $\mathbf{e}^\text{BERT}_i$ from the last layer of BERT.
This feature vector is fed into Bi-LSTM, generating two recurrent states $\mathbf{r}_i^\mathrm{f}$ and $\mathbf{r}_i^\mathrm{b}$:
\begin{align*}	
\mathbf{r}_i^\mathrm{f} &= {\mathrm{LSTM}^\text{forward}}\;\;\left( \mathbf{e}_i^\text{LEM} \oplus \mathbf{e}_i^\text{POS} \oplus \mathbf{e}^\text{BERT}_i \right)\\
\mathbf{r}_i^\mathrm{b} &= {\mathrm{LSTM}^\text{backward}}\left( \mathbf{e}_i^\text{LEM} \oplus \mathbf{e}_i^\text{POS} \oplus \mathbf{e}^\text{BERT}_i \right)
\end{align*}

\noindent Two multi-layer perceptrons (MLP) are then used to extract features for $w_i$ being a head $\mathbf{h}_i^\text{arc-h}$ or a dependent $\mathbf{h}_i^\text{arc-d}$, and two additional MLP are used to extract $\mathbf{h}_i^\text{rel-h}$ and $\mathbf{h}_i^\text{rel-d}$ for labeled parsing:
\begin{align*}
\mathbf{h}_i^\text{(arc-h)} &= \mathrm{MLP}^\text{(arc-h)}(\mathbf{r}_i^\mathrm{f} \oplus \mathbf{r}_i^\mathrm{b}) \in \mathbb{R}^{k \times 1}\\
\mathbf{h}_i^\text{(arc-d)} &= \mathrm{MLP}^\text{(arc-d)}(\mathbf{r}_i^\mathrm{f} \oplus \mathbf{r}_i^\mathrm{b}) \in \mathbb{R}^{k \times 1}\\
\mathbf{h}_i^\text{(rel-h)} &= \mathrm{MLP}^\text{(rel-h)}(\mathbf{r}_i^\mathrm{f} \oplus \mathbf{r}_i^\mathrm{b}) \in \mathbb{R}^{l \times 1}\\
\mathbf{h}_i^\text{(rel-d)} &= \mathrm{MLP}^\text{(rel-d)}(\mathbf{r}_i^\mathrm{f} \oplus \mathbf{r}_i^\mathrm{b}) \in \mathbb{R}^{l \times 1}
\end{align*}

\noindent $\mathbf{h}_{1..n}^\text{arc-h}$ are stacked into a matrix $H^\text{arc-h}$ with a bias for the prior probability of each token being a head, and $\mathbf{h}_{1..n}^\text{arc-d}$ are stacked into another matrix $H^\text{arc-d}$ as follows ($n$: \# of tokens, $U^{(arc)} \in \mathbb{R}^{k \times (k+1)}$):
\begin{align*}
H^\text{(arc-h)} &= (\mathbf{h}_1^\text{(arc-h)}, \ldots, \mathbf{h}_n^\text{(arc-h)}) \in \mathbb{R}^{k \times n} \\
H^\text{(arc-d)} &= (\mathbf{h}_1^\text{(arc-d)}, \ldots, \mathbf{h}_n^\text{(arc-d)}) \oplus \mathbf{1} \in \mathbb{R}^{(k+1) \times n} \\
S^\text{(arc)} &= H^{\text{(arc-h)}\top} \cdot U^{(\text{arc})} \cdot H^{\text{(arc-d)}} \in \mathbb{R}^{n \times n} 
\end{align*}

	
\noindent $S^\text{(arc)}$ is called a bilinear classifier that predicts head words.
Additionally, arc labels are predicted by another biaffine classifier $S^\text{(arc)}$, which combines $m$ bilinear classifiers for multi-classification ($m$: \# of labels, $U^\text{(rel)} \in \mathbb{R}^{l \times (l+1)}$, $V^\text{(rel)} \in \mathbb{R}^{(2\cdot l+1) \times m}$):
\begin{align*}
H^\text{(rel-h)} &= (\mathbf{h}_1^\text{(rel-h)}, \ldots, \mathbf{h}_n^\text{(rel-h)}) \in \mathbb{R}^{l \times n} \\
H^\text{(rel-d)} &= (\mathbf{h}_1^\text{(rel-d)}, \ldots, \mathbf{h}_n^\text{(rel-d)}) \oplus \mathbf{1} \in \mathbb{R}^{(l+1) \times n} \\
\mathcal{U}_i^\text{rel} &= H^{(\text{rel-h})\top} \cdot U_i^\text{(rel)} \cdot H^{(\text{rel-d})} \in \mathbb{R}^{n \times n} \\
\mathbf{S}^{\text{(rel)}} &= (\mathcal{U}_1^\text{rel}, \ldots, \mathcal{U}_m^\text{rel})\\
                 &+ (H^{(\text{rel-h})} \oplus H^{(\text{rel-d})})^\top \cdot V^{(\text{rel})} \in \mathbb{R}^{m \times n \times n}
\end{align*}

\noindent During training, softmax cross-entropy is used to optimize $S^\text{(arc)}$ and $S^\text{(rel)}$. Note that for the optimization of $S^\text{(rel)}$, gold heads are used instead of predicted ones.
During decoding, a maximum spanning tree algorithm is adopted for searching the optimal tree based on the scores in $S^\text{(arc)}$.


\subsection{Biaffine Attention for Semantic Parsing}
\label{ssec:bert-biaffine-graph}

\citet{dozat:18a} adapted their original biaffine parser to generate dependency graphs for semantic parsing, where each token can have zero to many heads.
Since the tree structure is no longer guaranteed, sigmoid cross-entropy is used instead so that independent binary predictions can be made for every token to be considered a head of any other token.
The label predictions are made as outputting the labels with the highest scores in $S^\text{(rel)}$ once arc predictions are made, as illustrated in Figure~\ref{fig:biaffine}.

This updated implementation is further simplified in our approach by removing the trainable word embeddings, the character-level feature detector, and their corresponding linear transformers.
Moreover, instead of using the interpolation between the head and label losses, equal weights are applied to both losses, reducing hyperparameters to tune.


\cleardoublepage
\section{Experiments}
\label{sec:experiments}

Three sets of experiments are conducted to evaluate the impact of our approaches using BERT (Sec.~\ref{sec:approach}).
For sequence tagging (Section~\ref{ssec:sequence-tagging}), part-of-speech tagging is chosen where each token gets assigned with a fine-grained POS tag.
For syntactic parsing (Section~\ref{ssec:syntactic-parsing}), dependency parsing is chosen where each token finds exactly one head, generating a tree per sentence.
For semantic parsing (Section~\ref{ssec:semantic-parsing}), semantic dependency parsing is chosen where each token finds zero to many heads, generating a graph per sentence.
Every task is tested on both English and Chinese to ensure robustness across languages.

Standard datasets are adapted to all experiments for fair comparisons to many previous approaches.
All our models are experimented three times and average scores with standard deviations are reported.
Section~\ref{sec:supplemental-materials} describes our environmental settings and data split in details for the replication of this work.


\subsection{Sequence Tagging}
\label{ssec:sequence-tagging}

For part-of-speech tagging, the Wall Street Journal corpus from the Penn Treebank 3~\cite{marcus:93a} is used for English, and the Penn Chinese Treebank 5.1~\cite{xue:05a} is used for Chinese.
Table~\ref{tbl:sequence-tagging-results} shows tagging results on the test sets.

\begin{table}[htbp!]
\centering\small

\subfigure[Results from the English test set. BERT$_{\tt BS}$ and BERT$_{\tt LG}$ are BERT's uncased base and cased large models, respectively.]{
\resizebox{\columnwidth}{!}{
\begin{tabular}{l||l|l}
 & \multicolumn{1}{c|}{\bf ALL} & \multicolumn{1}{c}{\bf OOV} \\
\hline\hline
\citet{ma:16a}                & 97.55             & \multicolumn{1}{c}{\bf 93.45} \\
\citet{ling:15a}              & 97.78             & \multicolumn{1}{c}{n/a} \\
\citet{clark:18a}             & 97.79             & \multicolumn{1}{c}{n/a} \\ 
\citet{akbik:18a}             & 97.85 ($\pm$0.01) & \multicolumn{1}{c}{n/a} \\
\citet{bohnet:18a}            & \bf 97.96         & \multicolumn{1}{c}{n/a} \\
\hline
Baseline                      & 97.70 ($\pm$0.05) & 92.44 ($\pm$0.03) \\
Baseline \BS\ BERT$_{\tt BS}$ & 96.96 ($\pm$0.06) & 91.23 ($\pm$0.22) \\
Baseline \BS\ BERT$_{\tt LG}$ & 96.96 ($\pm$0.05) & 91.26 ($\pm$0.25) \\
Baseline + BERT$_{\tt BS}$    & 97.68 ($\pm$0.06) & 92.69 ($\pm$0.32) \\
Baseline + BERT$_{\tt LG}$    & 97.67 ($\pm$0.02) & 93.01 ($\pm$0.27) \\
\end{tabular}}}

\subfigure[Results from the Chinese test set. * are evaluated on the character-level due to automatic segmentation, so their results are not directly comparable to ours but reported for reference.]{
\resizebox{\columnwidth}{!}{
\begin{tabular}{l||l|l}
 & \multicolumn{1}{c|}{\bf ALL} & \multicolumn{1}{c}{\bf OOV} \\
\hline\hline
\citet{zhang:15a}  & 94.47$^*$ & \multicolumn{1}{c}{n/a} \\
\citet{zhang:14a}  & 94.62$^*$ & \multicolumn{1}{c}{n/a} \\
\citet{kurita:17a} & 94.84$^*$ & \multicolumn{1}{c}{n/a} \\
\hline
\citet{hatori:11a} & 94.64     & \multicolumn{1}{c}{n/a} \\
\citet{wang:14a}   & 96.0      & \multicolumn{1}{c}{n/a} \\
\hline\hline
Baseline           & 95.65 ($\pm$0.26)          & 83.57 ($\pm$0.55) \\
Baseline \BS\ BERT & 96.38 ($\pm$0.15)          & 88.13 ($\pm$0.72) \\
Baseline + BERT    & \textbf{97.25} ($\pm$0.18) & \textbf{90.53} ($\pm$0.91)\\
\end{tabular}}}

\caption{Test results for part-of-speech tagging, where token-level accuracy is used as the evaluation metric. ALL: all tokens, OOV: out-of-vocabulary tokens.}
\label{tbl:sequence-tagging-results}
\end{table}

\noindent For English, the baseline is our replication of the Flair model using both GloVe and Flair embeddings (Section~\ref{ssec:bert-bilstm-crf}).
It shows a slightly lower accuracy, -0.15\%, than the original model~\cite{akbik:18a} due to the lack of variational dropout.
\BS BERT substitutes GloVe and Flair with BERT embeddings, and +BERT uses all three types of embeddings.
The baseline outperforms all BERT models for the ALL test, implying that Flair's Bi-LSTM character language model is more effective than BERT's word-piece approach.
No significant difference is found between BERT$_{\tt BS}$ and BERT$_{\tt LG}$.
However, an interesting trend is found in the OOV test, where the +BERT$_{\tt LG}$ model shows good improvement over the baseline.
This implies that BERT embeddings can still contribute to the Flair model for OOV although the CNN character language model from \citet{ma:16a} is marginally more effective than +BERT for out-of-vocabulary tokens.

For Chinese, the Bi-LSTM-CRF model with FastText embeddings is used for baseline (Sec.~\ref{ssec:bert-bilstm-crf}).
\BS BERT that substitutes FastText embeddings with BERT and +BERT that adds BERT embeddings to the baseline show progressive improvement over the prior model for both the ALL and OOV tests.
+BERT gives an accuracy that is 1.25\% higher than the previous state-of-the-art using joint-learning between tagging and parsing~\cite{wang:14a}.


\subsection{Syntactic Parsing}
\label{ssec:syntactic-parsing}

The same datasets used for POS tagging, the Penn Treebank and the Penn Chinese Treebank (Section~\ref{ssec:sequence-tagging}), are used for dependency parsing as well.
Table~\ref{tbl:syntactic-parsing-results} shows parsing results on the test sets.

\begin{table}[htbp!]
\centering\small

\subfigure[Results from the English test set.]{
\resizebox{\columnwidth}{!}{
\begin{tabular}{l||l|l}
 & \multicolumn{1}{c|}{\bf UAS} & \multicolumn{1}{c}{\bf LAS} \\
\hline\hline
\citet{dozat:17a}   & 95.74 & 94.08 \\
\citet{kuncoro:17a} & 95.8  & 94.6  \\
\citet{ma:18a}      & 95.87 & 94.19 \\
\citet{choe:16a}    & 95.9  & 94.1  \\
\citet{clark:18a}   & 96.6	& 95.0  \\
\hline
Baseline            & 95.78 ($\pm$0.04)          & 94.04 ($\pm$0.04) \\
Baseline \BS\ BERT  & 96.76 ($\pm$0.09)          & 95.27 ($\pm$0.13) \\
Baseline + BERT     & \textbf{96.79} ($\pm$0.08) & \textbf{95.29} ($\pm$0.12) \\
\end{tabular}}}

\subfigure[Results from the Chinese test set.]{
\resizebox{\columnwidth}{!}{
\begin{tabular}{l||l|l}
 & \multicolumn{1}{c|}{\bf UAS} & \multicolumn{1}{c}{\bf LAS} \\
\hline\hline
\citet{dozat:17a}   & 89.30 & 88.23 \\
\citet{ma:18a}      & 90.59 & 89.29 \\
\hline\hline
Baseline            & 91.02 ($\pm$0.10) & 89.89 ($\pm$0.09) \\
Baseline \BS\ BERT  & 93.21 ($\pm$0.06) & 92.21 ($\pm$0.05) \\
Baseline + BERT     & \textbf{93.34} ($\pm$0.21) & \textbf{92.29} ($\pm$0.22) \\
\end{tabular}}}

\caption{Test results for dependency parsing, where unlabeled and labeled attachment scores (UAS and LAS) are used as the evaluation metrics.}
\label{tbl:syntactic-parsing-results}
\end{table}

\noindent Our simplified version of the biaffine parser (Section~\ref{ssec:bert-biaffine}) is used for baseline, where GloVe and FastText embeddings are used for English and Chinese, respectively.
The baseline model gives a comparable result to the original model~\cite{dozat:17a} for English, yet shows a notably better result for Chinese, which can be due to higher quality embeddings from FastText.
\BS BERT substitutes the pre-trained embeddings with BERT and +BERT adds BERT embeddings to the baseline.
Moreover, BERT's uncased base model is used for English. 

Between \BS BERT and +BERT, no significant difference is found, implying that those pre-trained embeddings are not so useful when coupled with BERT.
All BERT models show significant improvement over the baselines for both languages, and outperform the previous state-of-the-art approaches using cross-view training~\cite{clark:18a} and stack-pointer networks~\cite{ma:18a} by 0.29\% and 3\% in LAS for English and Chinese, respectively.
Considering the simplicity of our +BERT models, these results are remarkable.


\vspace{-0.5ex}
\subsection{Semantic Parsing}
\label{ssec:semantic-parsing}

The English dataset from the SemEval 2015 Task 18: Broad-Coverage Semantic Dependency Parsing~\cite{oepen:15a} and the Chinese dataset from the SemEval 2016 Task 9: Chinese Semantic Dependency Parsing~\cite{che:16a} are used for semantic dependency parsing.

\vspace{-0.3ex}
\begin{table}[htbp!]
\centering\small

\subfigure[Results from the in-domain (ID) test sets.]
{\resizebox{\columnwidth}{!}{
\begin{tabular}{l||c|c|c||c}
 & \bf DM & \bf PAS & \bf PSD & \bf AVG \\
\hline\hline
\citet{du:15a}      & 89.1 & 91.3 & 75.7 & 85.3 \\
\citet{almeida:15a} & 89.4 & 91.7 & 77.6 & 86.2 \\
\citet{wang:18a}    & 90.3 & 91.7 & 78.6 & 86.9 \\
\citet{peng:17a}    & 90.4 & 92.7 & 78.5 & 87.2 \\  
\citet{dozat:18a}   & 93.7 & 93.9 & 81.0 & 89.5 \\
\hline\hline
Baseline            &     92.48  &     94.56  &     85.00  &     90.68  \\
Baseline \BS\ BERT  &     94.37  &     96.03  &     86.59  &     92.33  \\
Baseline + BERT     & \bf 94.58  & \bf 96.13  & \bf 86.80  & \bf 92.50  \\
\end{tabular}}}

\subfigure[Results from the out-of-domain (OOD) test sets.]
{\resizebox{\columnwidth}{!}{
\begin{tabular}{l||c|c|c||c}
 & \bf DM & \bf PAS & \bf PSD & \bf AVG \\
\hline\hline
\citet{du:15a}      & 81.8 & 87.2 & 73.3 & 80.8 \\
\citet{almeida:15a} & 83.8 & 87.6 & 76.2 & 82.5 \\
\citet{wang:18a}    & 84.9 & 87.6 & 75.9 & 82.8 \\
\citet{peng:17a}    & 85.3 & 89.0 & 76.4 & 83.6 \\  
\citet{dozat:18a}   & 88.9 & 90.6 & 79.4 & 86.3 \\
\hline\hline
Baseline            &     86.98  &     91.35  &     77.28  &     85.34  \\
Baseline \BS\ BERT  &     90.49  &     94.31  &     79.31  &     88.07  \\
Baseline + BERT     & \bf 90.86  & \bf 94.38  & \bf 79.48  & \bf 88.21  \\
\end{tabular}}}

\caption{Test results for semantic dependency parsing in English; labeled dependency F1 scores are used as the evaluation metrics. The standard deviations are reported in Section~\ref{ssec:sup-semantic-parsing}. DM: DELPH-IN dependencies, PAS: Enju dependencies, PSD: Prague dependencies, AVG: macro-average of (DM, PAS, PSD).}
\label{tbl:english-semantic-parsing-results}
\end{table}

\noindent Table~\ref{tbl:english-semantic-parsing-results} shows the English results on the test sets.
The baseline, \BS BERT, and +BERT models are similar to the ones in Section~\ref{ssec:syntactic-parsing}, except they use the sigmoid instead of the softmax function in the output layer to accept multiple heads (Section~\ref{ssec:bert-biaffine-graph}).
Our baseline is a simplified version of \citet{dozat:18a}; its average scores are 1.2\% higher and 1.0\% lower than the original model for ID and OOD, due to different hyperparameter settings.
+BERT shows good improvement over \BS BERT for both test sets, implying that BERT embeddings are complementary to those pre-trained embeddings, and surpasses the previous state-of-the-art scores by 3\% and 2\% for ID and OOD, respectively.

\begin{table}[htbp!]
\centering
\centering\resizebox{\columnwidth}{!}{
\begin{tabular}{l||c|c||c|c}
 & \multicolumn{2}{c||}{\bf NEWS} & \multicolumn{2}{c}{\bf TEXT} \\
\cline{2-5} & \bf UF & \bf LF & \bf UF & \bf LF \\
\hline\hline
\citet{artsymenia:16a} &     77.64 &     59.06 &     82.41 &     68.59 \\
\citet{wang:18a}       &     81.14 &     63.30 &     85.71 &     72.92 \\
\hline\hline
Baseline               &     80.51  &     64.90  &     88.06  &     77.28  \\
Baseline \BS\ BERT     &     82.91  &     67.17  &     90.83  & \bf 80.46  \\
Baseline + BERT        & \bf 82.92  & \bf 67.27  & \bf 91.10  &     80.41  \\
\end{tabular}}

\caption{Test results for semantic dependency parsing in Chinese, where unlabeled and labeled dependency F1 scores (UF and LF) are used as the evaluation metrics. The standard deviations are also reported in Section~\ref{ssec:sup-semantic-parsing}. NEWS: newswire, TEXT: textbook.}
\label{tbl:chinese-semantic-parsing-results}
\end{table}

\noindent Table~\ref{tbl:chinese-semantic-parsing-results} shows the Chinese results on the test sets.
No significant difference is found between \BS BERT and +BERT.
+BERT significantly outperforms the previous state-of-the-art by 4\% and 7.5\%  in LF for NEWS and TEXT, which confirms that BERT embeddings are very effective for semantic dependency parsing in both English and Chinese.

\begin{figure*}[htbp!]
\centering

\subfigure[English: Flair]
{\centering\FBOX{\includegraphics[width=0.48\columnwidth]{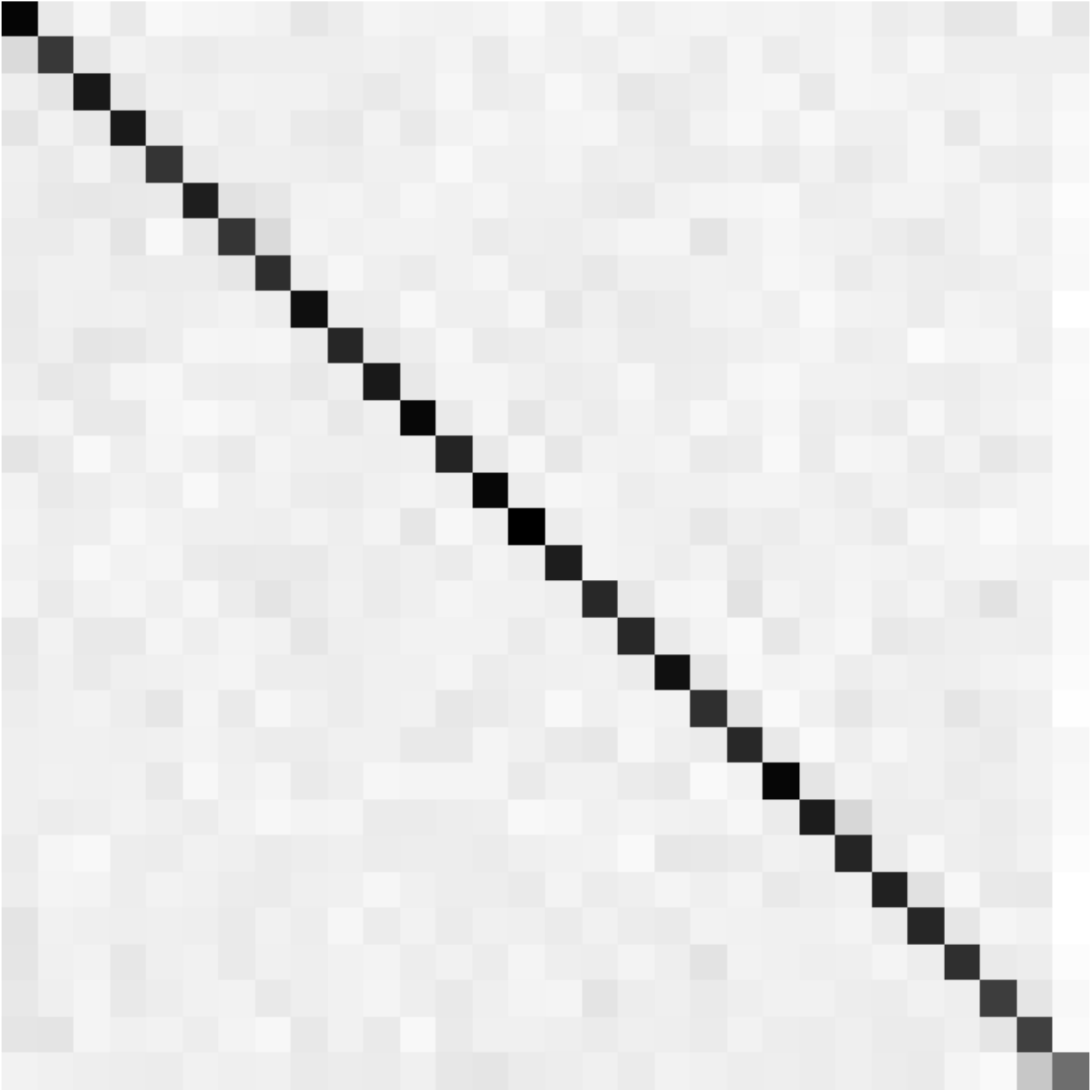}}}
~
\subfigure[English: BERT]
{\centering\FBOX{\includegraphics[width=0.48\columnwidth]{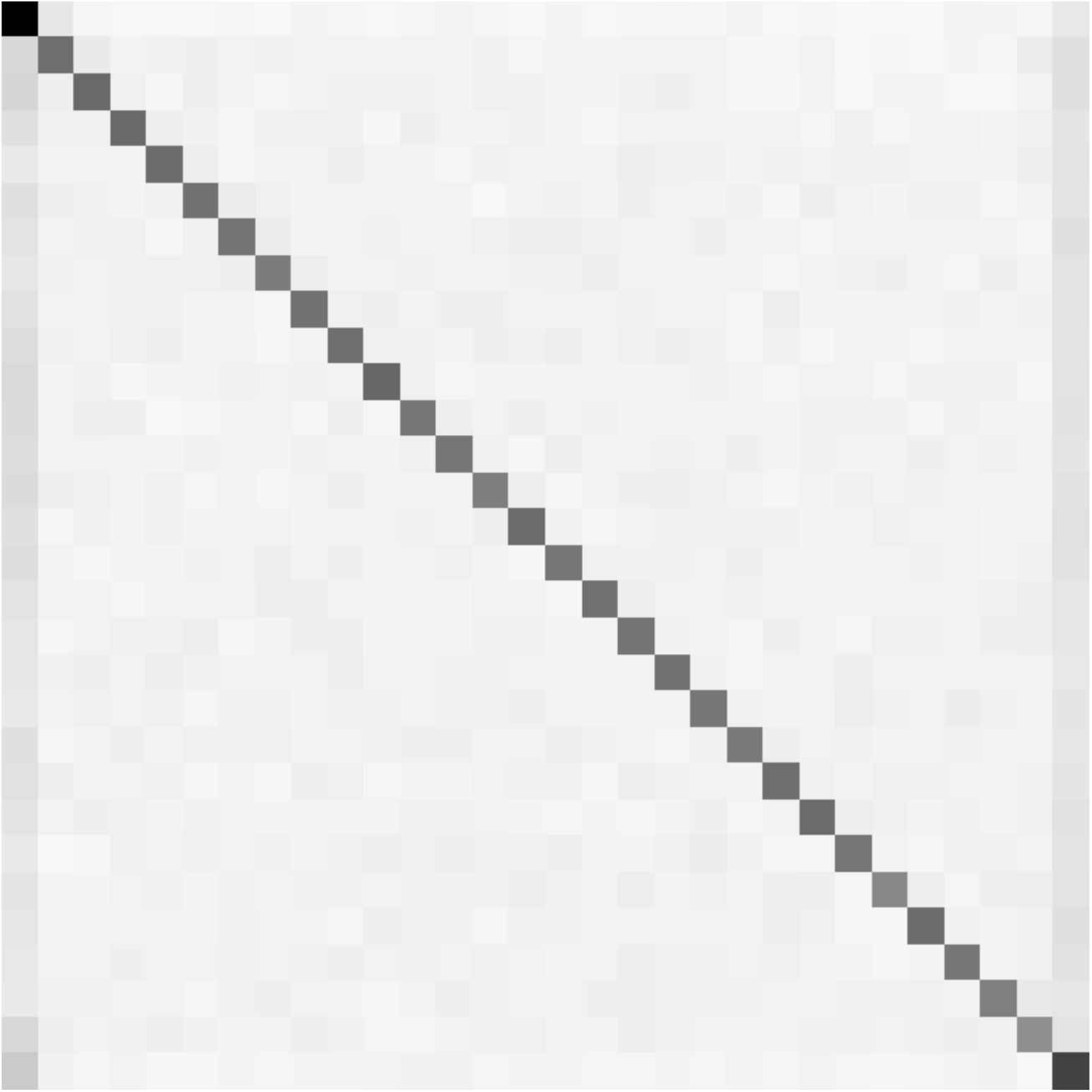}}}
~
\subfigure[Chinese: FastText]
{\centering\FBOX{\includegraphics[width=0.48\columnwidth]{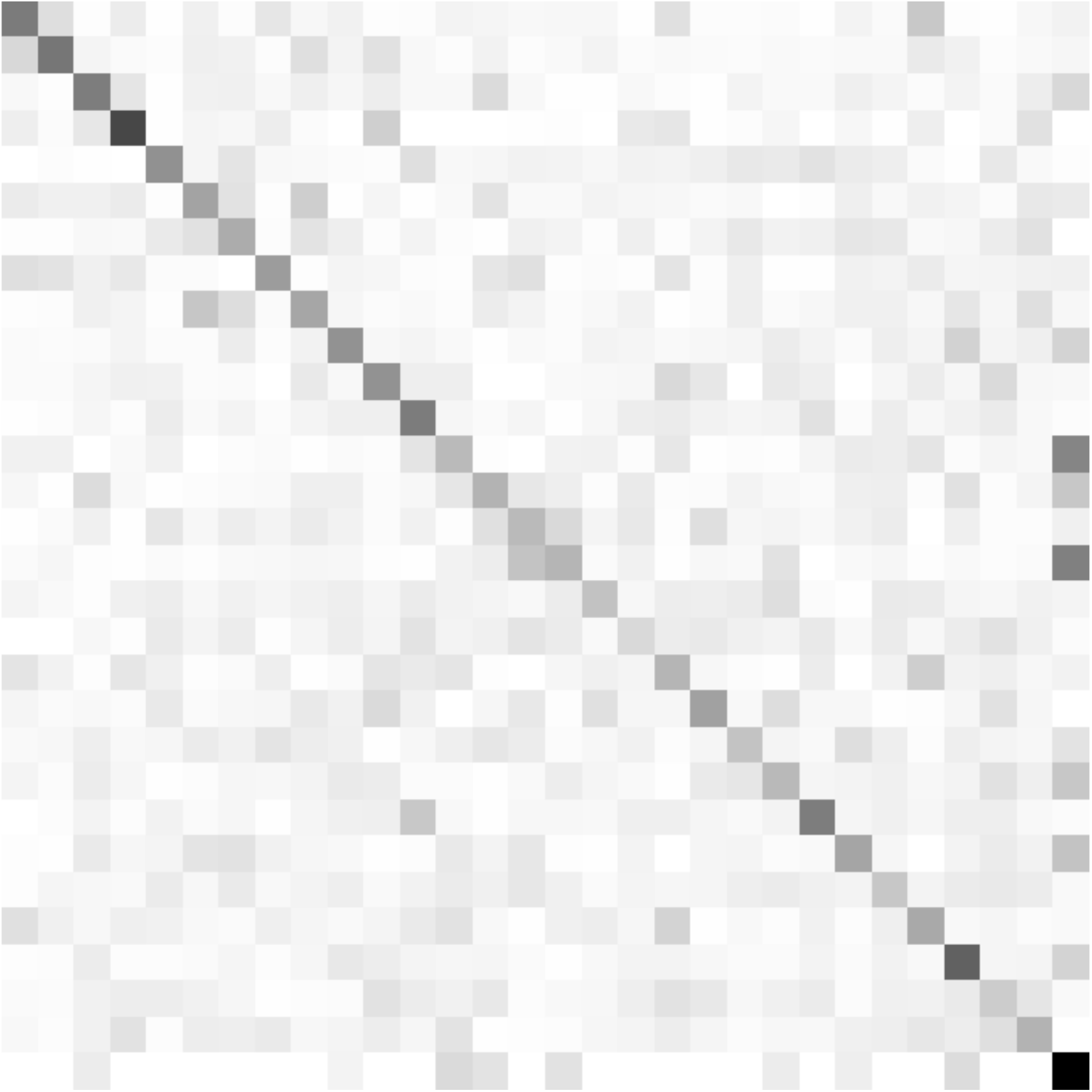}}}
~
\subfigure[Chinese: BERT]
{\centering\FBOX{\includegraphics[width=0.48\columnwidth]{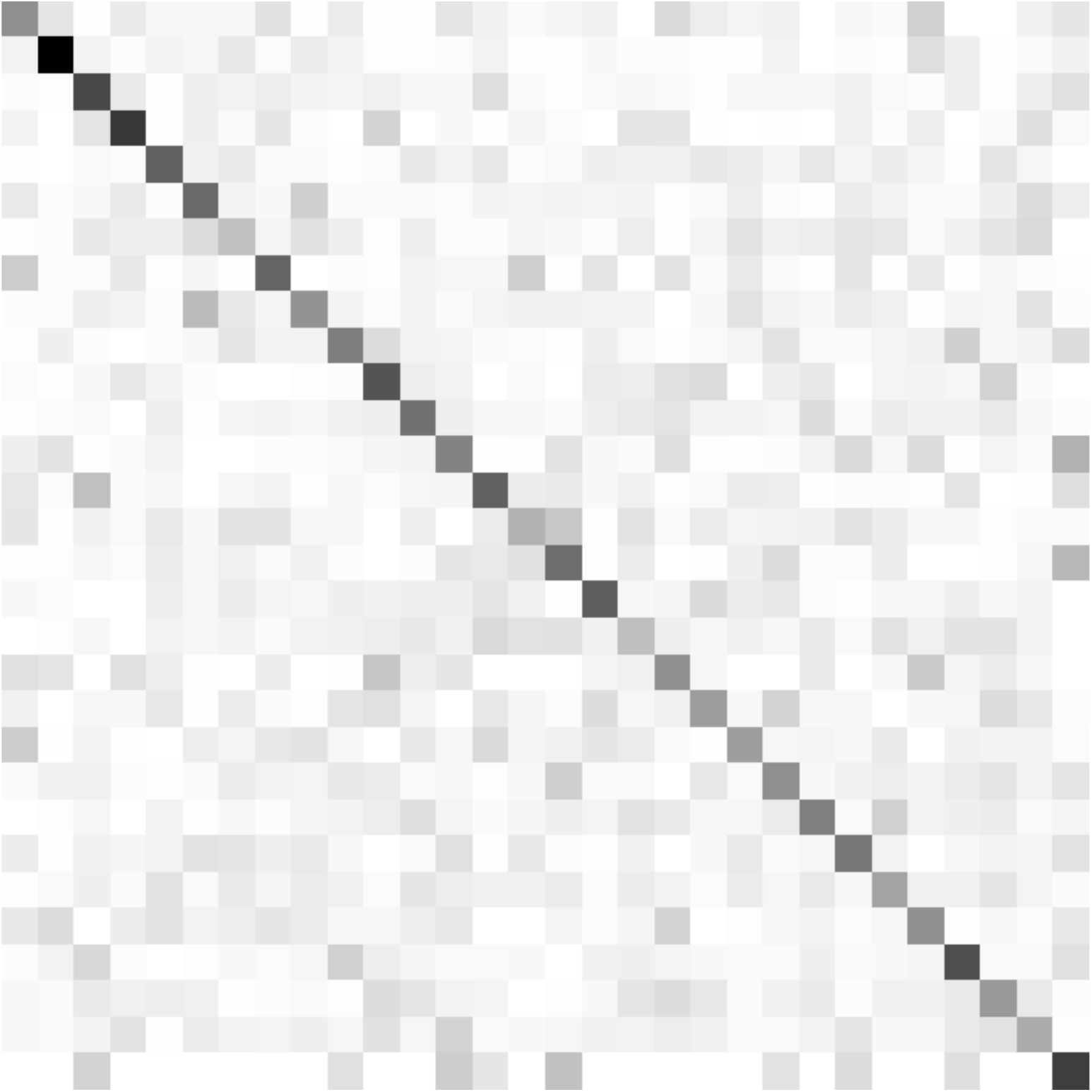}}}

\caption{Averaged attention matrices on sentences with 30 tokens. Each cell depicts the attention weight between $w_i$ and $w_j$, representing $i$'th and $j$'th tokens. All models are based on the Bi-LSTM-CRF~\cite{huang:15a} using the Flair~\cite{akbik:18a}, FastText~\cite{bojanowski:17a}, and BERT~\cite{devlin:18a} embeddings.}
\label{att-weight}
\end{figure*}

\section{Analysis}
\label{sec:analysis}

This section gives an in-depth analysis of the great results achieved by our approaches (Section~\ref{sec:experiments}) to better understand the role of BERT in these tasks.


\subsection{Attention Analysis for Tagging}
\label{ssec:attention-analysis}

The performance of \BS BERT models is surprisingly low for English POS tagging, compared to even a linear model achieving the accuracy of 97.64\% on the same dataset~\cite{choi:16a}.
This aligns with the findings reported by BERT~\cite{devlin:18a} and ELMo~\cite{peters:18a}, another popular contextualized embedding approach, where their POS and named entity tagging results do not surpass the state-of-the-art.
To study how tagging models are trained with BERT embeddings, we augment the baseline and \BS BERT$_{\tt BS}$ models in Table~\ref{tbl:sequence-tagging-results}(a) with dot-product self-attention~\cite{luong:15a}, and extract their attention weights.
We then average the attention matrices decoded from sentences with an equal length, 30 tokens, to find any general trend.

Comparing attention matrices across languages, it is clear that the Chinese matrices are much more checkered, implying that it requires more contents to make correct predictions in Chinese than English.
This makes sense because Chinese words tend to be more polysemous than English ones~\cite{huang:07a} so that they rely more on contents to disambiguate their categories.
For the Flair and BERT models in English, the Flair matrix is more checkered and its diagonal is darker, implying that it uses more contents while individual token embeddings convey more information for POS tagging so their weights are higher than the ones in the BERT matrix.
For the FastText and BERT models in Chinese, on the other hand, the BERT model is slightly more checkered and its diagonal is darker, indicating that BERT is better suited for this task than FastText.

\begin{figure}[htbp!]
\centering

\subfigure[Chinese: FastText]
{\centering\includegraphics[width=0.48\columnwidth]{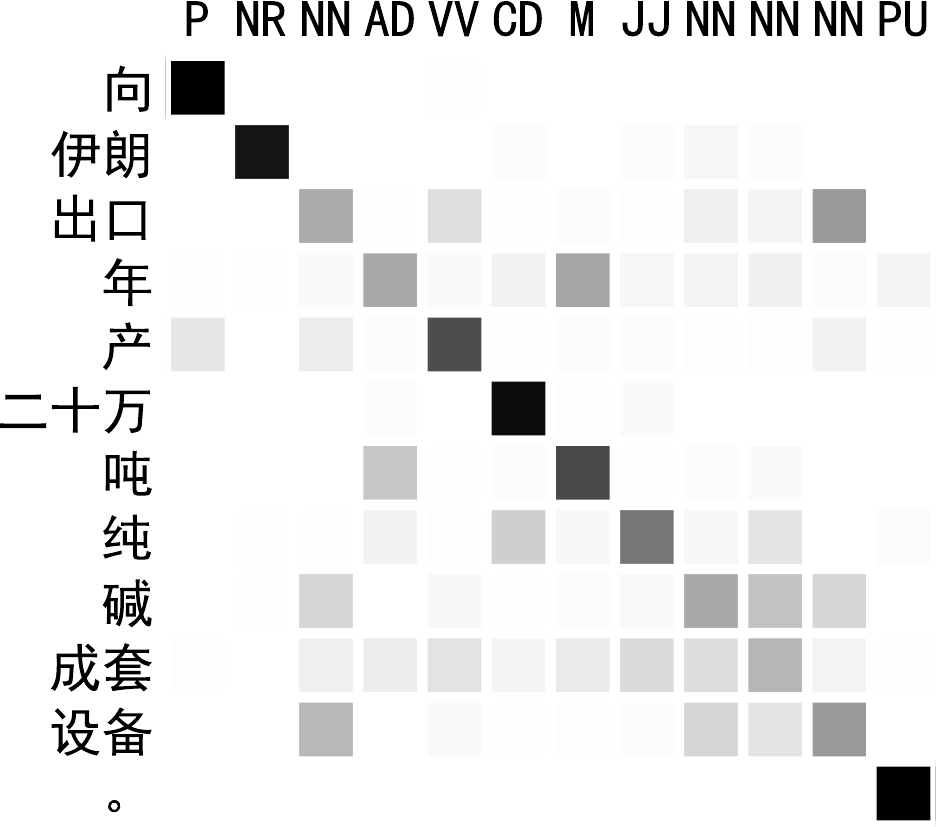}}
~
\subfigure[Chinese: BERT]
{\centering\includegraphics[width=0.48\columnwidth]{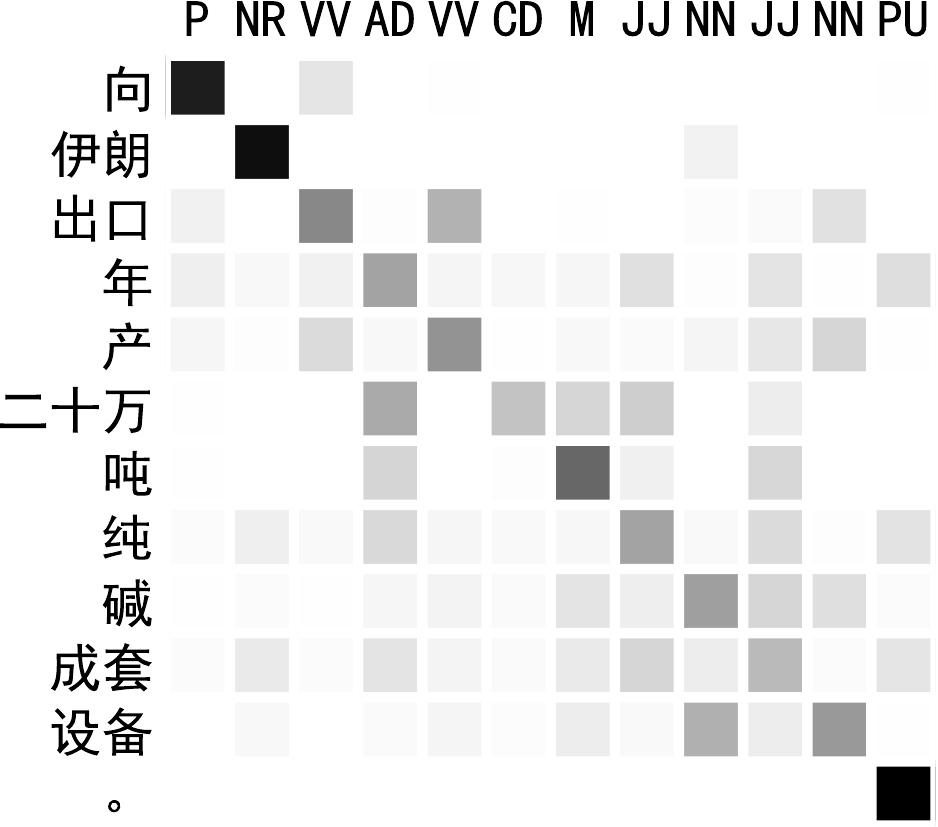}}

\caption{Attention matrices for the Chinese sentence: \CN{向\textsubscript{to} 伊朗\textsubscript{Iran} 出口\textsubscript{export} 年\textsubscript{yearly} 产\textsubscript{produce} 二十万\textsubscript{200K} 吨\textsubscript{tons} 纯\textsubscript{pure} 碱\textsubscript{alkali} 成套\textsubscript{whole} 设备\textsubscript{equipment} 。}, that can be translated to ``(China) export(s) (the) whole (set of) equipment(s) to Iran (that) yearly produce 200K tons (of) pure alkali''. X-axis: predicted POS tags.} 
\label{fig:att-sample}
\end{figure}

\noindent Figure~\ref{fig:att-sample} shows the attention matrices from a sample Chinese sentence.
The FastText model mispredicts \CN{出口\textsubscript{export}} and \CN{成套\textsubscript{whole}} as nouns, whereas the BERT model correctly predicts them as a verb and an adjective, respectively.
Notice that the BERT model gives the highest attention to \CN{产\textsubscript{produce}} for tagging \CN{出口\textsubscript{export}}, which both happen to be verbs, whereas the Flair model gives the highest attention to \CN{设备\textsubscript{equipment}} that is a noun.

\subsection{Far-distance Analysis for Parsing}
\label{ssec:distance-analysis}

\begin{figure*}[htbp!]
\centering

\subfigure[ID: DM]
{\centering\includegraphics[width=0.15\textwidth]{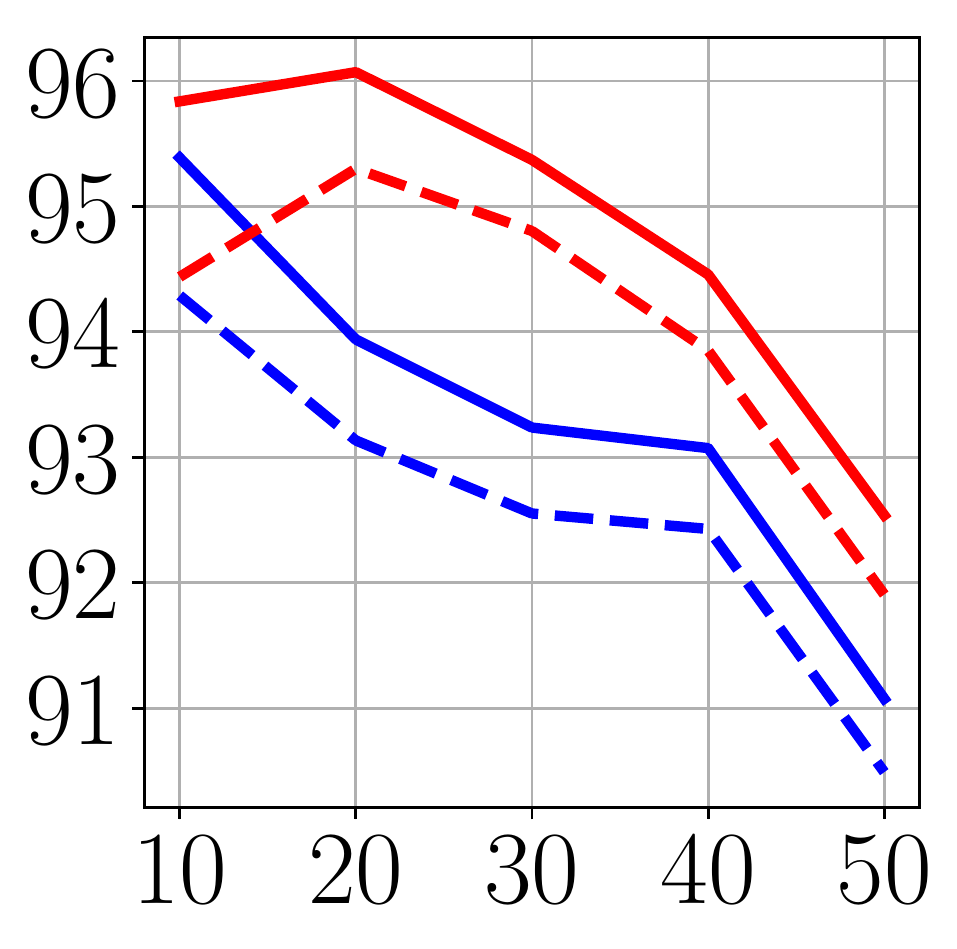}}
~
\subfigure[ID: PAS]
{\centering\includegraphics[width=0.15\textwidth]{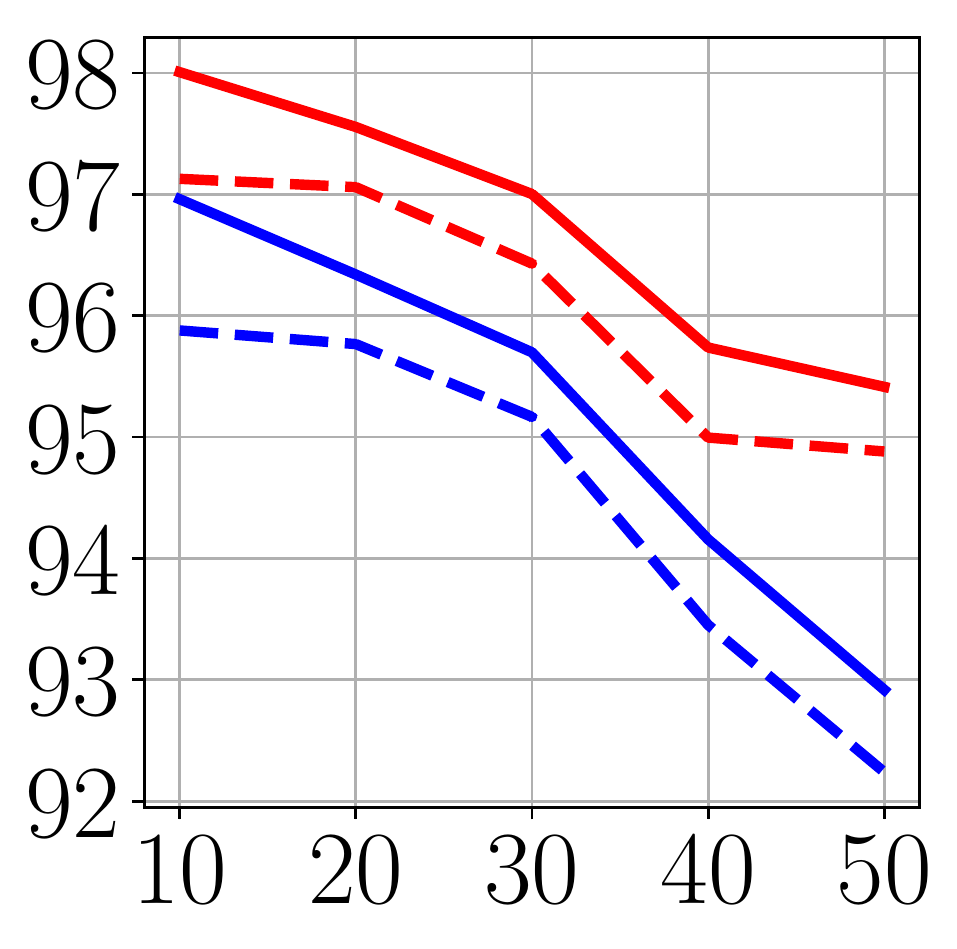}}
~
\subfigure[ID: PSD]
{\centering\includegraphics[width=0.15\textwidth]{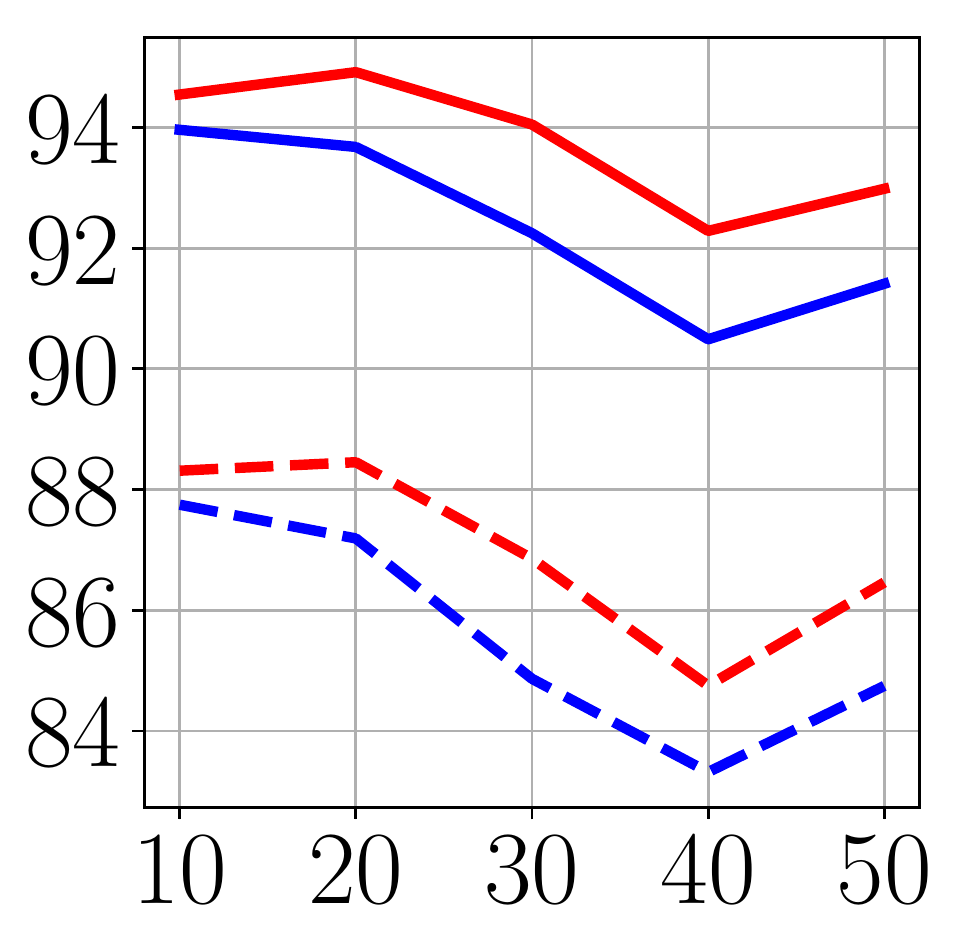}}
~
\subfigure[OOD: DM]
{\centering\includegraphics[width=0.15\textwidth]{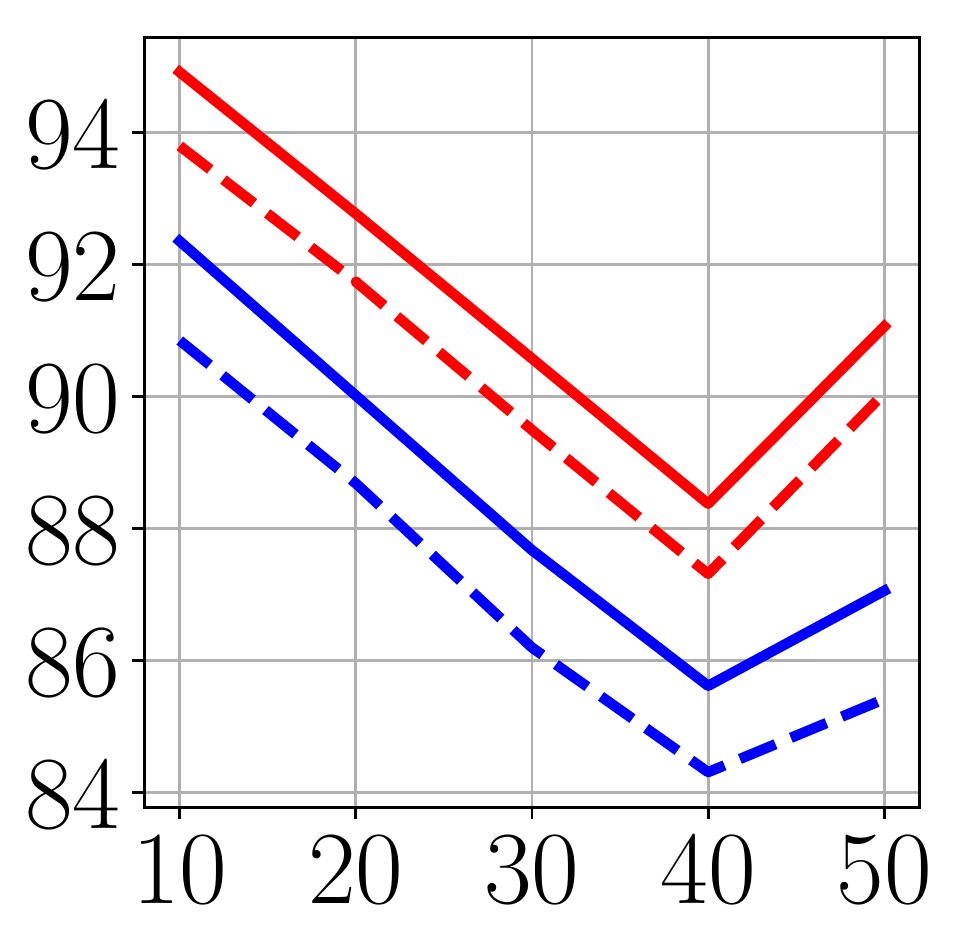}}
~
\subfigure[OOD: PAS]
{\centering\includegraphics[width=0.15\textwidth]{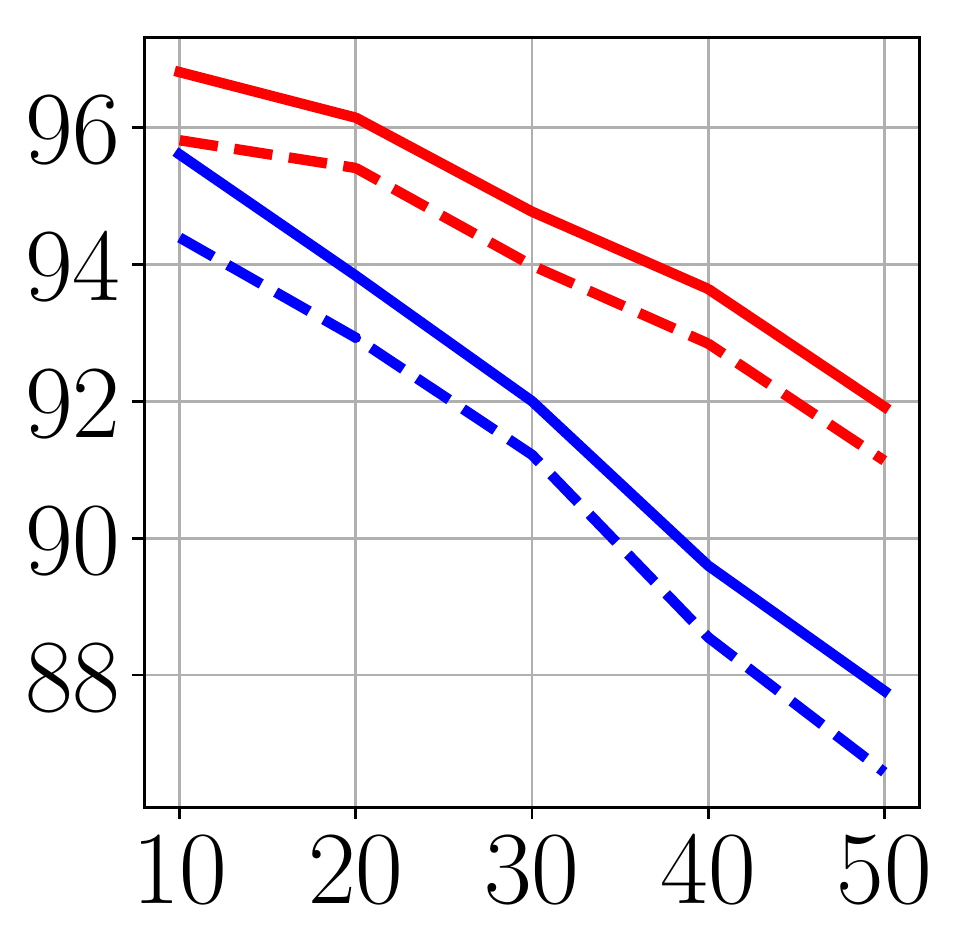}}
~
\subfigure[OOD: PSD]
{\centering\includegraphics[width=0.15\textwidth]{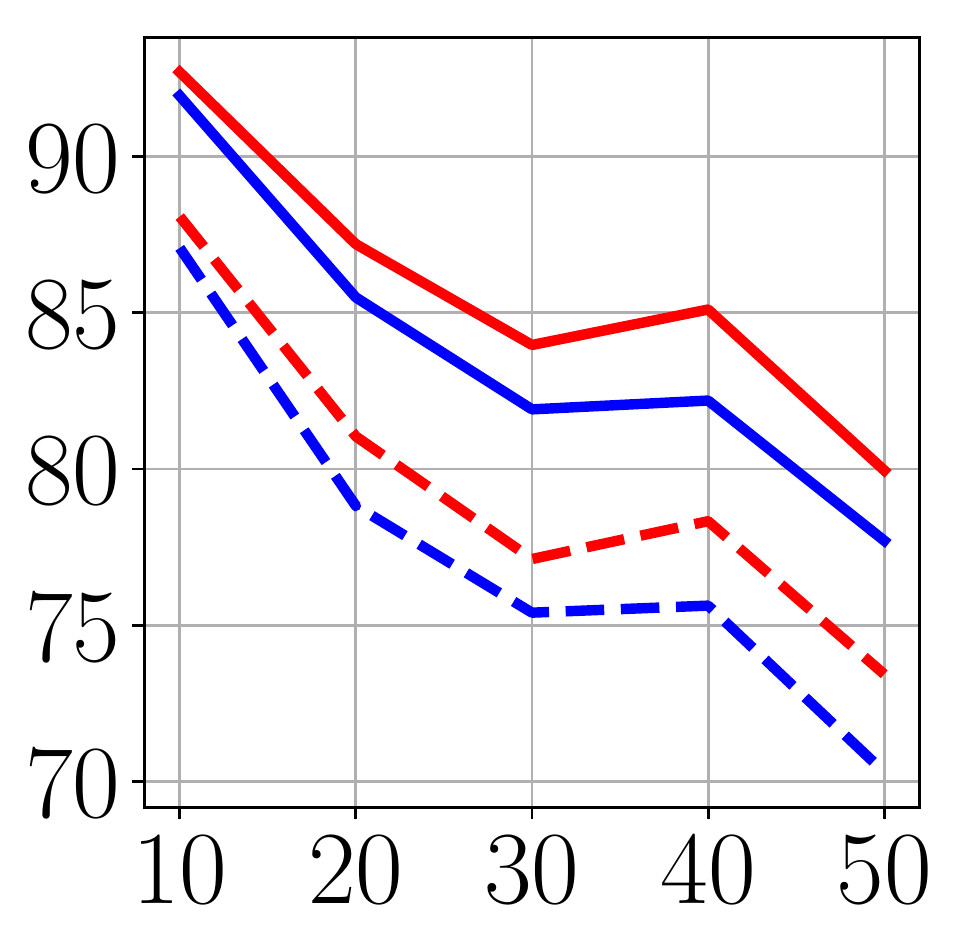}}
\caption{Average labeled F1 scores for semantic parsing w.r.t.\ sentence lengths. LAS and UAS are represented by solid and dashed lines, and scores from the baseline and \BS BERT models are shown in blue and red, respectively.}
\label{fig:dep-long}
\end{figure*}

The outputs of the baseline and \BS BERT models on semantic dependency parsing (Table~\ref{tbl:english-semantic-parsing-results}) are further analyzed for its robustness on long sentences.
The average F1 scores for each sentence group, ranging 1-50, are displayed in Figure~\ref{fig:dep-long}.
For DM and PAS, the baseline scores drop faster than those of \BS BERT as sentences get longer.
For PSD, the score drop rates are similar between the two, due to the challenging nature of this dataset~\cite{oepen:15a}.
This reflects that BERT embeddings can handle far-distant dependencies in longer sentences better.

One possible explanation to BERT's high capability of handling long sentences more robustly is its training objective and structure of masked language modeling (MLM).
MLM is trained to predict randomly masked tokens through features extracted by a bidirectional Transformer, which takes up to 512 tokens as input.
This is about twice larger than what recurrent neural networks typically expect in practice before gradient vanishes~\cite{khandelwal:18a}, and an order of magnitude larger than the context windows used by FastText or GloVe.
As a result, BERT embeddings can carry information from much farther-distant tokens, leading to higher performance on tasks requiring contextual understanding such as parsing.

\begin{figure*}[htbp!]
\centering
\includegraphics[width=\textwidth]{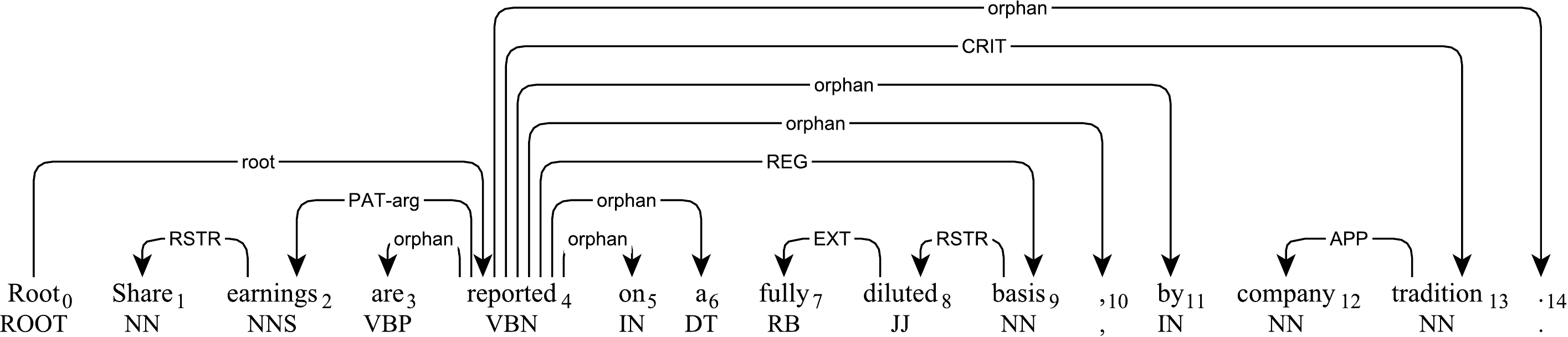}
\caption{"Share earnings are reported on a fully diluted basis , by company tradition ."}
\label{fig:deptree}
\end{figure*}


\subsection{Labeling Analysis on Semantic Parsing}
\label{ssec:label-analysis}

\begin{figure}[htbp!]
\centering

\subfigure[ID label ranking results. \{ID: \texttt{ID}, CR: \texttt{CRIT}, DI: \texttt{DIR2}, TT: \texttt{TTILL}, CO: \texttt{COND}\}, \{EX: \texttt{EXT-arg}, MA: \texttt{MANN-arg}, GR :\texttt{GRAD.member}, AI :\texttt{AIM-arg}, TW :\texttt{TWHEN-arg}\}.]
{\centering\includegraphics[width=\columnwidth]{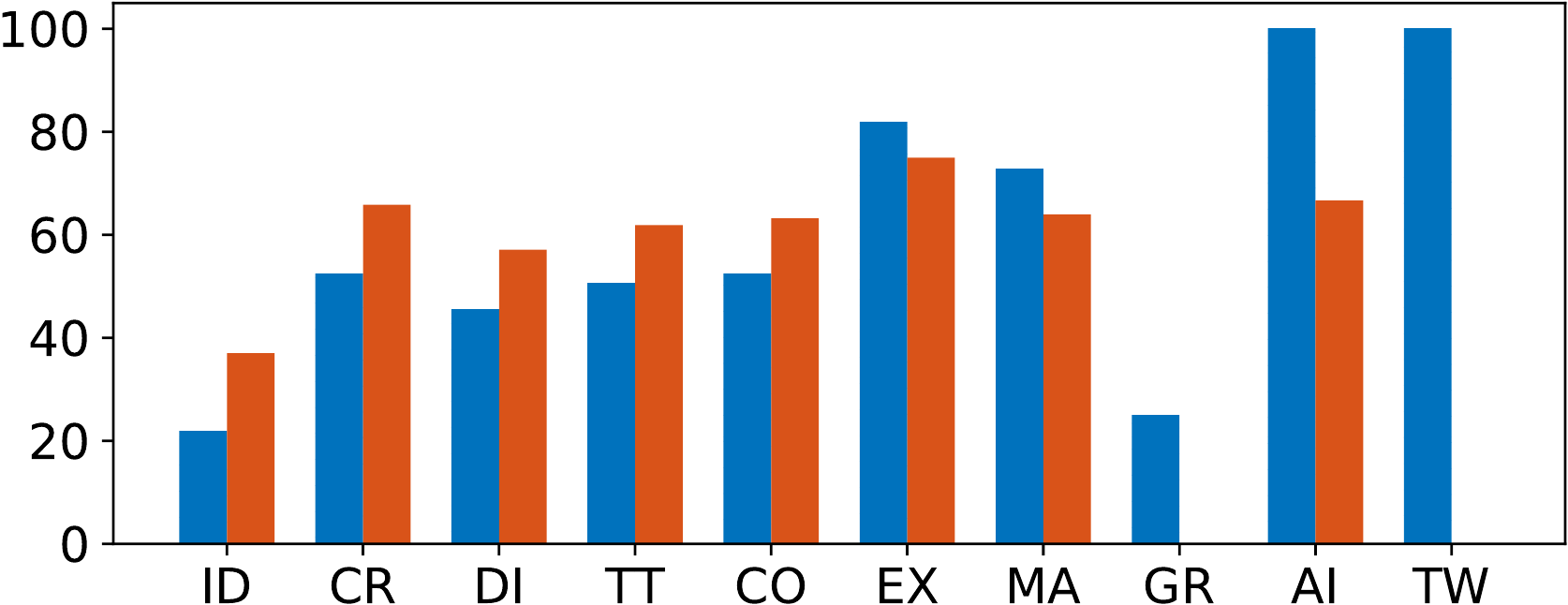}}
\vspace{-0.75ex}

\subfigure[OOD label ranking results. \{TF: \texttt{TFRWH}, TS: \texttt{TSIN}, RE: \texttt{RESTR}, SM: \texttt{SM}, PA: \texttt{PAR}\}, \{VO: \texttt{VOCAT}, LO: \texttt{LOC-arg}, AC: \texttt{ACMP-arg}, MA :\texttt{MANN-arg}, EX: \texttt{EXT-arg}\}.]
{\centering\includegraphics[width=\columnwidth]{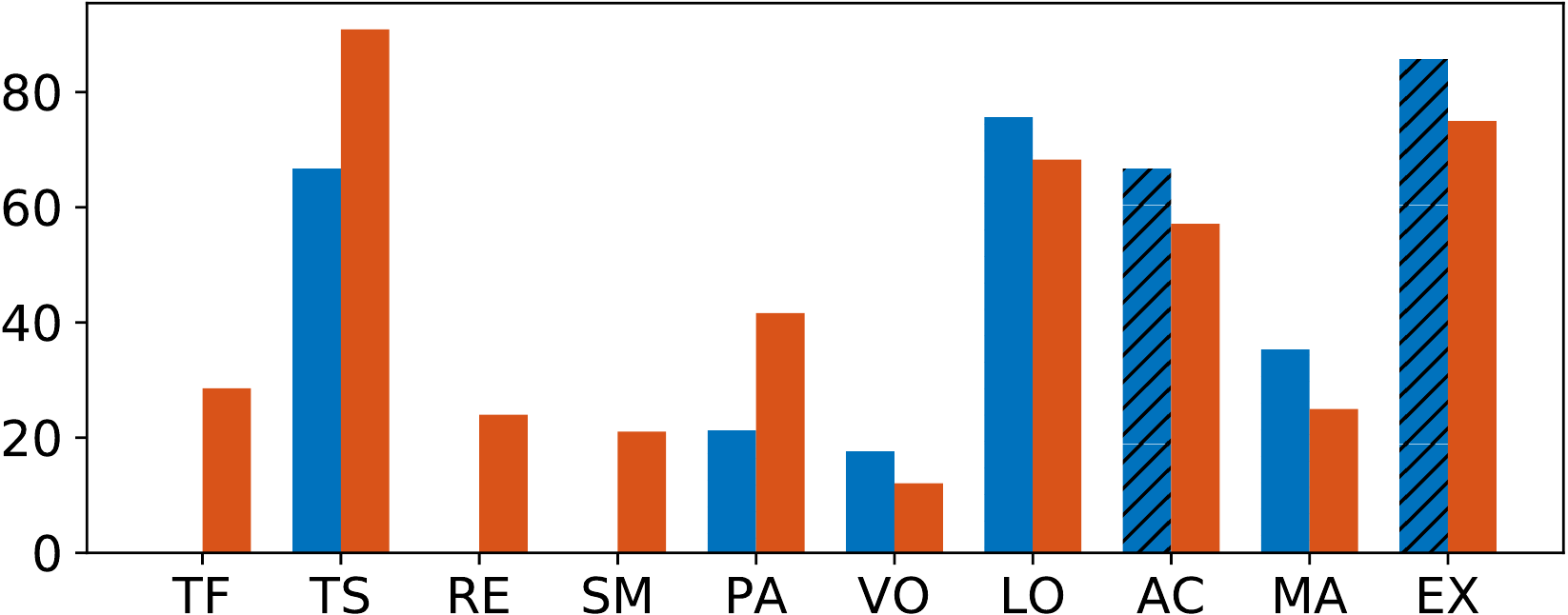}}
\vspace{-0.75ex}

\caption{Top-5 labels that \BS BERT outperforms the baseline (in red) and vice versa (in blue) on PSD.}
\label{fig:label-comp}
\end{figure}

Prague Semantic Dependencies (PSD) is used for our labeling analysis because it is manually annotated and well-documented~\cite{cinkova2006annotation}.
The average labeled F1 score of each label is ranked by the difference between the baseline and \BS BERT models in Table~\ref{tbl:english-semantic-parsing-results}.
Figure \ref{fig:label-comp} shows the top-5 labels on which \BS BERT outperforms, and vice versa.

\noindent The baseline performs better on certain arguments involving syntactic relations such as \texttt{LOC-arg} (locative), where the relation usually finds a preposition as the head of a noun phrase.
\BS BERT shows robust generalization for arguments involving semantic reasoning i.e., \texttt{CRIT} (criterion) or \texttt{COND} (condition).
For \textit{tradition}$_{13}$ in Figure~\ref{fig:deptree}, \BS BERT correctly classifies the \texttt{CRIT} label, while the baseline misclassified it as \texttt{ACT-arg} (argument of action).
The far-dependent relation between \textit{tradition}$_{13}$ and \textit{reported}$_{4}$ requires deeper inference on the context, which may be beyond the capacity of the baseline.

\section{Conclusion}
\label{sec:conclusion}

In this paper, we describe our methods of exploiting BERT as token-level embeddings for tagging and parsing tasks.
Our experiments empirically show that tagging and parsing can be tackled using much simpler models without losing accuracy.
Out of 12 datasets, our approaches with BERT have established new state-of-the-art for 11 of them. 
As the first work of employing BERT with syntactic and semantic parsing, our approach is much simpler yet more accurate than the previous state-of-the-art. 

Through a dedicated error analysis and extensive dissections based on an attention mechanism, we uncover interesting properties of BERT from syntactics, semantics, and multilingual perspectives. 
Beyond syntactically intensive or morphologically complex tasks, BERT embeddings are well-suited for semantic reasoning in long sentences. 
\bibliography{emnlp-ijcnlp-2019}
\bibliographystyle{acl_natbib}
\cleardoublepage\appendix
\section{Supplemental Materials}
\label{sec:supplemental-materials}

Throughout this paper, we use the following notations for data splits,
\texttt{TRN}: training, \texttt{DEV}: development, \texttt{TST}: test.

\subsection{Part-of-Speech Tagging}

For English, the Wall Street Journal corpus from the Penn Treebank 3~\cite{marcus:93a} is used with the standard split for part-of-speech tagging.
The baseline is our replication of the Flair model~\cite{akbik:18a} using embeddings trained by GloVe. Specifically, we use $100$-dim GloVe embeddings\footnote{\url{http://nlp.stanford.edu/data/glove.6B.zip}} \cite{pennington2014glove} trained on Wikipedia 2014 and Gigaword 5 involving 6B tokens in total. 

\begin{table}[htbp!]
\centering\small
\begin{tabular}{c||r|r|r}
\bf Set & \multicolumn{1}{c|}{\bf Sections} & \multicolumn{1}{c|}{\bf Sentences} & \multicolumn{1}{c}{\bf Tokens} \\
\hline\hline
\texttt{TRN} &  0-18 & 38,219 & 912,344 \\
\texttt{DEV} & 19-21 &  5,527 & 131,768 \\
\texttt{TST} & 22-24 &  5,462 & 129,654 \\
\end{tabular}
\caption{English part-of-speech tagging on Penn Treebank 3~\cite{marcus:93a}.}
\label{tbl:stat-ptb-pos}
\end{table}

\noindent For Chinese, the Penn Chinese Treebank 5.1~\cite{xue:05a} is used with the standard split for POS.
The baseline is our replication of the Bi-LSTM-CRF model~\cite{huang:15a}. We use $300$-dim FastText\footnote{\url{https://fasttext.cc/docs/en/crawl-vectors.html}} with subword information. 

\begin{table}[htbp!]
\centering\small
\begin{tabular}{c||r|r|r}
\bf Set & \multicolumn{1}{c|}{\bf Sections} & \multicolumn{1}{c|}{\bf Sentences} & \multicolumn{1}{c}{\bf Tokens} \\
\hline\hline
\texttt{TRN} & 1-270; 400-1151 & 18,078 & 493,691 \\
\texttt{DEV} & 301-325         &    350 &   6,821 \\
\texttt{TST} & 271-300         &    348 &   8,008 \\
\end{tabular}
\caption{Chinese part-of-speech tagging on Penn Chinese Treebank 5.1~\cite{xue:05a}.}
\label{tbl:stat-ctb-pos}
\end{table}

%
%
%

\subsection{Syntactic Parsing}
\label{ssec:sup-syntatic-parsing}

For English, the Wall Street Journal corpus from the Penn Treebank 3 is used with the standard split, converted by the Stanford Parser 3.3.0\footnote{\url{http://nlp.stanford.edu/software/lex-parser.shtml}}, for syntactic dependency parsing.
For Chinese, the Penn Chinese Treebank 5.1 is used with the standard split, converted by the head-finding rules of \citet{zhang:08a} and the labeling rules of Penn2Malt\footnote{\url{https://cl.lingfil.uu.se/~nivre/research/Penn2Malt.html}}.
The POS tags are auto-generated by the POS tagger in NLP4J~\cite{choi:16a}\footnote{\url{https://emorynlp.github.io/nlp4j/}} using 10-way jackknifing on the training set for English, and the gold word segmentation and POS tags are used for Chinese.


\begin{table}[htbp!]
\centering\small
\begin{tabular}{c||r|r|r}
\bf Set & \multicolumn{1}{c|}{\bf Sections} & \multicolumn{1}{c|}{\bf Sentences} & \multicolumn{1}{c}{\bf Tokens} \\
\hline\hline
\texttt{TRN} & 2-21 & 39,832 & 950,027 \\
\texttt{DEV} &   22 &  1,700 &  40,117 \\
\texttt{TST} &   23 &  2,416 &  56,684 \\
\end{tabular}
\caption{English dependency parsing on Penn Treebank 3~\cite{marcus:93a}.}
\label{tbl:stat-ptb-dep}
\end{table}

\begin{table}[htbp!]
\centering\small
\begin{tabular}{c||r|r|r}
\bf Set & \multicolumn{1}{c|}{\bf Sections} & \multicolumn{1}{c|}{\bf Sentences} & \multicolumn{1}{c}{\bf Tokens} \\
\hline\hline
\texttt{TRN} &   1-815; 1001-1136 & 16,091 & 437,991 \\
\texttt{DEV} & 886-931; 1148-1151 &    803 &  20,454 \\
\texttt{TST} & 816-885; 1137-1147 &  1,910 &  50,319 \\
\end{tabular}
\caption{Chinese dependency parsing on Penn Chinese Treebank 5.1~\cite{xue:05a}.}
\label{tbl:stat-ctb-dep}
\end{table}

\subsection{Semantic Dependency Parsing}
\label{ssec:sup-semantic-parsing}

For English, the English dataset from the SemEval 2015 Task 18~\cite{oepen:15a} is used for semantic dependency parsing.
For Chinese, the SemEval 2016 Task 9~\cite{che:16a} dataset is used. However, SemEval 2015 Chinese dataset is not used because it is less popular.
The POS tags provided in those datasets are used as they are for both English and Chinese, and the provided word segmentation is used for Chinese.

\begin{table}[htbp!]
\centering\small

\subfigure[General statistics.]
{\parbox{\columnwidth}{\centering
\begin{tabular}{l||r|r|r}
\multicolumn{1}{c||}{\bf Set} & \multicolumn{1}{c|}{\bf Sections} & \multicolumn{1}{c|}{\bf Sentences} & \multicolumn{1}{c}{\bf Tokens} \\
\hline\hline
\texttt{TRN}     &  0-19 & 33,964 & 765,025 \\
\texttt{DEV}     &    20 &  1,692 &  37,692 \\
\texttt{TST-IN}  &    21 &  1,410 &  31,948 \\
\texttt{TST-OOD} & Brown &  1,849 &  31,583 \\
\end{tabular}}}

\subfigure[Number of dependencies.]
{\parbox{\columnwidth}{\centering
\begin{tabular}{l||r|r|r}
\multicolumn{1}{c||}{\bf Set} & \multicolumn{1}{c|}{\bf DM} & \multicolumn{1}{c|}{\bf PAS} & \multicolumn{1}{c}{\bf PSD} \\
\hline\hline
\texttt{TRN}     & 585,646 & 711,064 & 532,271 \\
\texttt{DEV}     &  28,854 &  34,775 &  26,438 \\
\texttt{TST-IN}  &  24,307 &  29,776 &  22,063 \\
\texttt{TST-OOD} &  23,497 &  28,569 &  20,095 \\
\end{tabular}}}

\caption{English semantic dependency parsing on SemEval 2015 Task 18: Broad-Coverage Semantic Dependency Parsing~\cite{oepen:15a}. Section 20 is used as development set, following the recommended split from the organizers.}
\label{tbl:stat-semeval-sdp}
\end{table}

\noindent The standard deviation from English and Chinese semantic parsing test sets are recorded in Table~\ref{tbl:english-semantic-parsing-std} and Table~\ref{tbl:chinese-semantic-parsing-std} respectively.

\begin{table}[htbp!]
\centering\small

\subfigure[Standard deviations from the in-domain (ID) test sets.]
{\resizebox{\columnwidth}{!}{
\begin{tabular}{l||c|c|c||c}
 & \bf DM & \bf PAS & \bf PSD & \bf AVG \\
\hline\hline
Baseline            &     ± 0.03 &      ± 0.03 &      ± 0.10 &      ± 0.04 \\
Baseline \BS\ BERT  &     ± 0.05 &      ± 0.04 &      ± 0.11 &      ± 0.05 \\
Baseline + BERT     &     ± 0.05 &      ± 0.01 &      ± 0.17 &      ± 0.06 \\
\end{tabular}}}

\subfigure[Standard deviations from the out-of-domain test sets.]
{\resizebox{\columnwidth}{!}{
\begin{tabular}{l||c|c|c||c}
 & \bf DM & \bf PAS & \bf PSD & \bf AVG \\
\hline\hline
Baseline            &      ± 0.10 &      ± 0.09 &      ± 0.12 &      ± 0.04 \\
Baseline \BS\ BERT  &      ± 0.16 &      ± 0.07 &      ± 0.18 &      ± 0.08 \\
Baseline + BERT     &      ± 0.15 &      ± 0.06 &      ± 0.01 &      ± 0.07 \\
\end{tabular}}}

\caption{Standard deviations for semantic dependency parsing in English; labeled dependency F1 scores are reported in Section~\ref{ssec:semantic-parsing}. DM: DELPH-IN dependencies, PAS: Enju dependencies, PSD: Prague dependencies, AVG: macro-average of (DM, PAS, PSD).}
\label{tbl:english-semantic-parsing-std}
\end{table}

\begin{table}[htbp!]
\centering
\centering\resizebox{\columnwidth}{!}{
\begin{tabular}{l||c|c||c|c}
 & \multicolumn{2}{c||}{\bf NEWS} & \multicolumn{2}{c}{\bf TEXT} \\
\cline{2-5} & \bf UF & \bf LF & \bf UF & \bf LF \\
\hline\hline
Baseline               &      ± 0.24 &      ± 0.13 &      ± 0.25 &      ± 0.81 \\
Baseline \BS\ BERT     &      ± 0.07 &      ± 0.04 &      ± 0.19 &      ± 0.84 \\
Baseline + BERT        &      ± 0.13 &      ± 0.09 &      ± 0.15 &      ± 0.10 \\
\end{tabular}}

\caption{Standard deviations for semantic dependency parsing in Chinese; average unlabeled and labeled dependency F1 scores (UF and LF) are reported in Section~\ref{ssec:semantic-parsing}. NEWS: newswire, TEXT: textbook.}
\label{tbl:chinese-semantic-parsing-std}
\end{table}

%
%

\subsection{Implementation}

Our models are implemented in MXNet, and ran on NVIDIA Tesla V100 GPUs. Note that in our implementation, the BERT large cased model requires 15GB GPU memory, which exceeds the memory limit of TITAN X (12GB). The training time for baseline+BERT models on each dataset is listed in Table \ref{tbl:trn-time}.

\begin{table}[htbp!]
\centering\small
\begin{tabular}{c|c}
\bf Dataset & \bf Time (hours)  \\ 
\hline\hline
  WSJ-POS     &  15     \\ 
\hline
    CTB-POS     &  11     \\ 
\hline
    PTB    &   6    \\ 
\hline
    CTB    &   10    \\ 
\hline
  SemEval 2015      &  6     \\ 
\hline
    SemEval 2016    &   10    \\ 
\end{tabular}
  \caption{Training time on specific datasets.}
  \label{tbl:trn-time}
\end{table}

\subsection{Hyperparameter}

\subsubsection{Tagging}

The hyperparameter configuration for English and Chinese tagging models are given in Table \ref{hyper-tagger}

\begin{table}[htbp!]
  \centering\small
  \begin{tabular}{l|l}
    \multicolumn{2}{c}{\bf Hidden Sizes}\\
    \hline \hline
    GloVe / FastText & 100 / 300\\
    Flair BiLSTM & 1 @ 2048\\
    BiLSTM & 1 @ 256\\
    BERT EN / CN & 1024 / 768\\
    \multicolumn{2}{c}{\bf Dropout Rates}\\
    \hline \hline
    Embedding(s) & 50\%\\
    \multicolumn{2}{c}{\bf Loss \&{} Optimizer}\\
    \hline \hline
    Optimizer & SGD\\
    Learning rate & $0.1$\\
    Anneal factor & $0.5$\\
    Anneal patience & $2$\\
    Batch size & $32$\\
    Max epochs & 150
  \end{tabular}
  \caption{Hyperparameter configuration for tagging.}
  \label{hyper-tagger}
\end{table}

\subsubsection{Parsing}

We have similar configurations for both syntactic and semantic dependency parsing in English and Chinese, shown in Table \ref{hyper-parser}.

\begin{table}[htbp!]
  \centering\small
  \begin{tabular}{l|l}
    \multicolumn{2}{c}{\bf Hidden Sizes}\\
    \hline \hline
    GloVe / FastText & 100 / 300\\
    $\bf{e}^{\text{lemma}}$ & 100\\
    $\bf{e}^{\text{tag}}$ & 100\\
    BiLSTM & 3 @ 400\\
    BERT EN / CN & 768 / 768\\
    $\text{MLP}^\text{(arc)}$ & 500\\
    $\text{MLP}^\text{(label)}$ & 100\\
    \multicolumn{2}{c}{\bf Dropout Rates}\\
    \hline \hline
    Embeddings & 33\%\\
    Word Dropout & 33\%\\
    Variational Dropout & 33\%\\
    MLP & 33\%\\
    \multicolumn{2}{c}{\bf Loss \&{} Optimizer}\\
    \hline \hline
    Optimizer & SGD\\
    Learning rate& $1e^{-3}$\\
    Adam $\beta_1$ & 0.9\\
    Adam $\beta_2$ & 0.9\\
    Adam $\epsilon$ & $1e^{-12}$\\
    Anneal factor & $0.75$\\
    Anneal every  & 5000\\
    Batch size & 5000\\
    Train steps & 50000
  \end{tabular}
  \caption{Hyperparameter configuration for parsing.}
  \label{hyper-parser}
\end{table}

\end{document}